\documentclass[twocolumn,letterpaper]{IEEEtran}
\usepackage{lineno,hyperref}
\usepackage{amsmath,amssymb,amsfonts}
\usepackage{amsthm}
\usepackage{algorithmic}
\usepackage{graphicx}
\usepackage{textcomp}
\usepackage{xcolor}
\usepackage{times}
\usepackage{epsfig}
\usepackage{graphicx}
\usepackage{algorithm}
\usepackage{algorithmic}
\usepackage{multirow}
\usepackage{booktabs}
\usepackage{array,color}
\usepackage{caption}
\usepackage{subfigure}
\usepackage{float}
\usepackage{supertabular}
\usepackage{longtable}
\usepackage{setspace}

\newtheorem{definition}{Definition}
\newtheorem{theorem}{Theorem}
\newtheorem{lemma}{Lemma}

\newtheorem{assumption}{Assumption}

\begin{document}

\title{Differentially Private ADMM Algorithms for Machine Learning}

\author{Tao~Xu,
		   Fanhua~Shang,~\IEEEmembership{Senior Member,~IEEE,}
           Yuanyuan~Liu,~\IEEEmembership{Member,~IEEE,}\\
		   Hongying~Liu,~\IEEEmembership{Member,~IEEE,}
           Longjie Shen,
           and~Maoguo Gong,~\IEEEmembership{Senior Member,~IEEE}
\IEEEcompsocitemizethanks{\IEEEcompsocthanksitem T.\ Xu, F.\ Shang (Corresponding author), Y.\ Liu, H.\ Liu, and L.\ Shen are with the Key Laboratory of Intelligent Perception and Image Understanding of Ministry of Education, School of Artificial Intelligence, Xidian University, China. E-mails: \{fhshang, yyliu, hyliu\}@xidian.edu.cn, xidianxutao@gmail.com, s269524963@gmail.com.\protect
\IEEEcompsocthanksitem M.\ Gong is with the Key Laboratory of Intelligent Perception and Image Understanding of Ministry of Education, School of Electronic Engineering, Xidian University, China. E-mail: gong@ieee.org.}
\thanks{Manuscript received October 30, 2020.}}

\markboth{}{Shell \MakeLowercase{\textit{et al.}}: Bare Demo of IEEEtran.cls for Computer Society Journals}
\maketitle
%IEEE Transactions on Information Forensics and Security

\begin{abstract}
In this paper, we study efficient differentially private alternating direction methods of multipliers (ADMM) via gradient perturbation for many machine learning problems. For smooth convex loss functions with (non)-smooth regularization, we propose the first differentially private ADMM (DP-ADMM) algorithm with performance guarantee of $(\epsilon,\delta)$-differential privacy ($(\epsilon,\delta)$-DP). From the viewpoint of theoretical analysis, we use the Gaussian mechanism and the conversion relationship between R\'enyi Differential Privacy (RDP) and DP to perform a comprehensive privacy analysis for our algorithm. Then we establish a new criterion to prove the convergence of the proposed algorithms including DP-ADMM. We also give the utility analysis of our DP-ADMM. Moreover, we propose an accelerated DP-ADMM (DP-AccADMM) with the Nesterov's acceleration technique. Finally, we conduct numerical experiments on many real-world datasets to show the privacy-utility tradeoff of the two proposed algorithms, and all the comparative analysis shows that DP-AccADMM converges faster and has a better utility than DP-ADMM, when the privacy budget $\epsilon$ is larger than a threshold.
\end{abstract}


%It has a high level of privacy protection, and the level of privacy protection can be evaluated by a thorough mathematical study. This makes differential privacy very convincing.

\section{Introduction}
\IEEEPARstart{I}{n} the era of `Big Data', most people's lives will be presented on the internet in different types of data, and personal privacy may be leaked when data are released/shared. Data is the most valuable resource for research institutions and decision-making departments, which can greatly improve the service system of society, but meanwhile we must also protect the privacy of users who provide data often containing personal identifiable information and various confidential data. Generally, research institutions use these data sets to train machine learning models. Sometimes the models are made publicly available, which gives attackers an opportunity to obtain individual privacy. Thus, it is necessary to add privacy preservation techniques during training and learning process.

In recent years, privacy-preserving technologies are very popular. Differential privacy (DP) \cite{dwork2006calibrating} that can provide rigorous guarantees for individual or personal privacy by adding randomized noise has been extensively studied in the literature \cite{Dwork:dp,gong:dpml}. Therefore, whether in academia or industry, differential privacy is the most recognized privacy preserving technology. A lot of applications have been studied to be equipped with differential privacy in many fields, such as data mining, machine learning and deep learning. For instance, differentially private recommender systems have been extensively studied \cite{mcsherry2009differentially}\cite{machanavajjhala2011personalized}\cite{zhu2013differential}\cite{zhu2017privacy} to deal with the privacy leakage during collaborative filtering process. Besides, personalized online advertising \cite{lindell2011practical}, health data \cite{dankar2012application}, face recognition \cite{chamikara2020privacy}\cite{othman2014privacy}, network trace analysis \cite{mcsherry2010differentially} and search logs \cite{gotz2011publishing} have all been studied by utilizing differential privacy for privacy-preserving. Of course, real-world applications have also been widely developed by technology companies and government agencies, such as Google \cite{erlingsson2014rappor}, Apple \cite{applelearning}, Microsoft \cite{ding2017collecting}, and US Census Bureau \cite{abowd2018us}. Thus, we can see that differential privacy plays an important role in privacy-preserving nowadays.

%v, which also faces the privacy preserving problem

Empirical risk minimization (ERM) is a commonly used supervised learning problem. Let $D\!=\!(l_1, l_2, \ldots, l_n)$ be a dataset with $n$ samples, where $l_i \!\in\! \mathbb R^d$. Many machine learning problems such as ERM are formulated as the following minimization problem:
\begin{equation} \label{eq:finite_sum_problem}
\min_{x \in \mathbb R^d} \; \big\{F(x) := \frac{1}{n}\sum_{i=1}^n f_i(x, l_i) + g(x)\big\}
\end{equation}
where each $f_i:\mathbb R^d\!\rightarrow\! \mathbb R$ is a smooth convex loss function, and $g:\mathbb R^d \rightarrow \mathbb R$ is a simple convex (non)-smooth regularizer such as the $\ell_1$-norm or $\ell_2$-norm regularizer. Many differential privacy algorithms have been proposed to deal with ERM problems, such as DP-SGD \cite{abadi2016deep}, DP-SVRG \cite{Wang2018Differentially}, and DP-SRGD \cite{wang2019efficient}. Therefore, this paper mainly considers the generalized ERM problem with more complex regularizers (e.g., $g(x)\!=\!\lambda\|Ax\|_1$ with a given matrix $A$ and a regularized parameter $\lambda \!>\! 0$), such as graph-guided fused Lasso~\cite{kim:flasso}, generalized Lasso~\cite{tibshirani:glasso} and graph-guided support vector machine (SVM) \cite{ouyang2013stochastic}.

%$\lambda > 0$ is a regularization parameter, sparsity inducing 
The alternating direction method of multipliers (ADMM) is an efficient and popular optimization method for solving the generalized ERM problem. Although there are many research works on differential privacy ADMMs, most of them focus on differential privacy of distributed ADMMs such as \cite{huang2020differentially,huang2019dp,wang:admm}. In fact, distributed ADMMs are mostly suitable for federated learning instead of centralized problems. Therefore, only a few of them work on the centralized and stochastic ADMMs \cite{chen2020renyi} or objective perturbed centralized and deterministic ADMMs such as \cite{wang2019differential} to deal with centralized problems.

However, the gradients of stochastic ADMMs such as \cite{ouyang2013stochastic} have their own gradient noise due to random sampling \cite{johnson:svrg}. Thus, it is difficult to estimate the noise added into stochastic algorithms. Generally, stochastic differentially private algorithms need some methods to estimate their privacy loss \cite{abadi2016deep,dwork2010boosting}. But rough estimations may lead to bad utility of algorithms. As for the position where we add the noise to, actually, gradient perturbation is a better choice than output/objective perturbation for first-order algorithms \cite{yu2019gradient}. Firstly, at each iteration, gradient perturbation can release the noisy gradient so that the utility of the algorithm will not be affected. Secondly, objective perturbation often requires strong assumptions on the objective function, while gradient perturbation only needs to bound the sensitivity of gradients. It does not require strong assumptions on the objective. Moreover, as for DP-ERM problems, gradient perturbation often achieves better empirical utility than output/objective perturbations. Thus, it is very meaningful to study differentially private ADMM algorithms under gradient perturbation. Until now, there is no gradient perturbed differential privacy work on centralized and deterministic ADMM algorithms. Therefore, in this paper, we will focus on centralized and deterministic ADMMs and propose efficient gradient perturbed differential privacy ADMM algorithms for solving the generalized ERM problem (i.e., Problem (\ref{equ002}) in Section II).

ADMM algorithms can deal with more complicated ERM problems, especially the ERM problem with equality constraints, e.g., graph-guided logistic regression and graph-guided SVM problems. In this paper, we propose two efficient deterministic differential privacy ADMM algorithms that satisfy $(\epsilon,\delta)$-differential privacy (DP), namely differential privacy ADMM (DP-ADMM) algorithm and differential privacy accelerated ADMM (DP-AccADMM) algorithm. Our algorithms can be applied to many real-world applications, such as finance, medical treatment, Internet and transportation. Because we give a quantitative representation of privacy leakage, even if attackers own the largest background, they can not get individual privacy information. That is, under our privacy guarantees, our algorithms satisfy $(\epsilon,\delta)$-DP.

In the first proposed algorithm (i.e., DP-ADMM), we replace $f(x)$ by its first-order approximation to get the gradient term $\nabla\! f$, and then add Gaussian noise as gradient perturbation into it. Moreover, we give the privacy guarantee analysis to show the size of the noise variance added to the algorithm, which can ensure adequate security of our algorithm. In particular, we use the relationship between R\'enyi differential privacy (RDP) and $(\epsilon,\delta)$-DP to get the privacy guarantee of our algorithm. Utility preserving is one of the important indicators used to measure the utility of our algorithm that seeks to preserve data privacy while maintaining an acceptable level of utility. Therefore, we provide the utility bound for our algorithm, which indicates how good the model can be trained. Unlike distributed differential privacy ADMMs \cite{huang2019dp,wang:admm}, we define a new convergence criterion to analyze the convergence property of our algorithm and give our utility bound.

However, common ADMMs converge very slowly when approaching the optimal solution, so is DP-ADMM. A common solution is to introduce an acceleration technique to ADMMs. Like the Nesterov method \cite{nesterov:fast} and the momentum acceleration method \cite{qian1999momentum}, the Nesterov accelerated method is a well-known momentum acceleration method \cite{nesterov:co}, and has a faster convergence rate than traditional momentum acceleration methods \cite{ruder2016overview}. In particular, Goldstein et al.\ \cite{goldstein2014fast} proved that their accelerated ADMM with Nesterov acceleration has a convergence rate $O(1/T^2)$, while traditional non-accelerated ADMMs only have the convergence rate $O(1/T)$, where $T$ is the number of iterations.

Therefore, in the second proposed algorithm (i.e., DP-AccADMM), we use the Nesterov acceleration technique to accelerate our DP-ADMM. The convergence speed of non-accelerated ADMMs will slow down gradually with the increase of the iteration number especially when the dataset is huge, so is DP-ADMM. Therefore, we also propose a new accelerated DP-AccADMM algorithm. Then we conduct some experiments to compare the performance of DP-ADMM and DP-AccADMM. Moreover, we give some comparative analysis for our experiments, which shows that DP-AccADMM converges much faster than DP-ADMM and retains a good performance on testing data, when the privacy budget $\epsilon$ becomes bigger and reaches a threshold.

Our contributions of this paper are summarized as follows:
\begin{itemize}
\item  Based on deterministic ADMMs, we propose two efficient differentially private ADMM algorithms for solving (non)-smooth convex optimization problems with privacy protection. In the proposed algorithms, each subproblem is solved exactly in a closed-form expression by using a first-order approximation.
\item  We use the relationship between RDP and $(\epsilon,\delta)$-DP to get the privacy guarantees of our DP-ADMM algorithm. Moreover, we design a new convergence criterion to complete the convergence analysis of DP-ADMM. It is non-trivial to provide the utility bound and gradient complexity for our algorithm. To the best of our knowledge, we are the first to give the theoretical analysis of gradient perturbed differentially private ADMM algorithms in the centralized and deterministic settings.
\item  We empirically show the effectiveness of the proposed algorithms by performing extensive empirical evaluations on graph-guided fused Lasso models and comparing them with their counterparts. The results show that DP-AccADMM performs much better than DP-ADMM in terms of convergence speed. In particular, DP-AccADMM continuously improves performance on test sets with the increase of privacy budget $\epsilon$ and even outperforms DP-ADMM.
\end{itemize}

The remainder of this paper is organized as follows. Section~\ref{sec2} discusses recent research advances in differential privacy methods. Section~\ref{sec3} introduces the  preliminary of differential privacy. We propose two new efficient differential privacy ADMM algorithms and analyze their privacy guarantees in Section~\ref{sec4}. Experimental results in Section~\ref{sec5} show the effectiveness of our algorithms. In Section~\ref{sec6}, we conclude this paper and discuss future work.

\section{Related Work}
\label{sec2}
In this section, we present the formulation considered in this paper, and review some differential privacy methods.

%, and a closed convex set $\mathcal{C}\subseteq \mathbb{R}^p$,
\subsection{Problem Setting}
For solving the generalized ERM problem (\ref{eq:finite_sum_problem}) with a smooth convex function $f(x)\!=\!\frac{1}{n}\sum_{i=1}^{n}\!{f_i(x,l_i)}$ and a complex sparsity inducing regularizer, we introduce an auxiliary variable $y$ and an equality constraint $Ax+By=c$, and can reformulate Problem (\ref{eq:finite_sum_problem}) as follows:
\begin{align}
\label{equ002}
\!\!\!\min_{x\in\mathbb{R}^{d_{1}}\!,y\in\mathbb{R}^{d_{2}}}\!\! \big\{f(x)+g(y),\;\textup{s.t.,}\; Ax+By=c\big\}\!\!\!
\end{align}
where $A\in\mathbb{R}^{d_{3}\times d_{1}}$, $B\in\mathbb{R}^{d_{3}\times d_{2}}$, and $c\in\mathbb{R}^{d_{3}}$. Therefore, this paper aims to propose efficient differentially private ADMM algorithms for solving the more general equality constrained minimization problem (\ref{equ002}).

%u^T(Ax+By-c)
Let $\rho>0$ be a penalty parameter, and $u$ be a dual variable. Then the augmented Lagrangian function of Problem (\ref{equ002}) is:
\begin{align*}
	L_{\rho}(x,y,u) = &\,f(x)+g(y)+\langle u,\,Ax+By-c\rangle \\&+(\rho/2)\|Ax+By-c\|^2.
\end{align*}
In an alternating or sequential fasion, at iteration $t$, ADMM performs the following update rules:
\begin{eqnarray}
&\!\!y_t=\arg\min_y \big\{g(y)+\frac{\rho}{2}\| Ax_{t-1}+By-c+u_{t-1}\|^2\big\},\label{equ003}\\
&\!\!x_t=\arg\min_x \big\{f(x)+\frac{\rho}{2}\| Ax+By_t-c+u_{t-1}\|^2\big\},\;\;\;\label{equ004}\\
&\!\!u_t=u_{t-1}+Ax_t+By_t-c.\qquad\quad\qquad\qquad\qquad\;\:\:\;\:\;\;\,\,\label{equ005}
\end{eqnarray}
This is the classic update form of ADMM \cite{boyd2011distributed}. While updating the variable $x$, this update step usually has a high computational complexity, especially when the dataset is very large.

%\begin{equation}\label{equ003}
%y_t=\arg\min_y \Big\{g(y)+\frac{\rho}{2}\| Ax_{t-1}+By-c+u_{t-1}\|^2\Big\},
%\end{equation}
%\begin{equation}\label{equ004}
%x_t=\arg\min_x \Big\{f(x)+\frac{\rho}{2}\| Ax+By_t-c+u_{t-1}\|^2\Big\},
%\end{equation}
%\begin{equation}\label{equ005}
%u_t=u_{t-1}+Ax_t+By_t-c.
%\end{equation}

\subsection{Related Work}
As for differential privacy, there are three main types of perturbation used to solve the empirical risk minimization problems under differential privacy \cite{Dwork:dp,gong:dpml}. Output perturbation is to perturb the model parameters. For instance, \cite{Li2007t} analyzed that introducing output perturbation can make $k$-anonymous algorithms satisfy differential privacy. \cite{chaudhuri2011differentially} analyzed the sensitivity of optimal solution between neighboring databases. Objective perturbation is to perturb the objective function trained by algorithms. \cite{Fukuchi2017Differentially} proposed an algorithm introduced objective perturbation to solve ERM problems and analyzed its privacy guarantee to satisfy $(\epsilon,\delta)$-DP. Gradient perturbation is to perturb the gradients used for updating parameters by first-order optimization methods. For instance, \cite{Wang2018Differentially} proposed DP-SVRG by introducing gradient perturbation to the SVRG operator in \cite{johnson:svrg}, and used moment accountant to complete both its privacy guarantee and utility guarantee.

In recent years, there are some research works on the ADMM algorithms that satisfies differential privacy. However, most of them are about distributed ADMM algorithms. For instance, \cite{huang2019dp}  analyzed the relationship between privacy guarantee and utility guarantee on their differentially private distributed ADMM algorithms. Of course, there are few research works focused on differentially private centralized ADMM algorithms. For example, \cite{Chen2019Renyi} proposed two stochastic ADMM algorithms, which satisfy $(\alpha,\beta)$-RDP and provide privacy guarantee of the algorithms. However, there is no research work on differentially private centralized and deterministic ADMM algorithms. Therefore, in this paper, we focus on differentially private centralized deterministic ADMM algorithms, and propose two efficient deterministic DP-ADMM and DP-AccADMM algorithms with privacy guarantees.

\section{Preliminaries}
\label{sec3}
\textbf{Notations.} Throughout this paper, the norm $\|\!\cdot\!\|$ is the standard Euclidean norm, $\|\!\cdot\!\|_{1}$ denotes the $\ell_{1}$-norm, i.e., $\|x\|_{1}\!=\!\sum_{i}\!|x_{i}|$, and $\|\!\cdot\!\|_{2}$ is the spectral norm (i.e., the largest singular value of the matrix). We denote by $\nabla\!f(x)$ the gradient of $f(x)$ if it is differentiable, or $g'(x)$ any of the subgradients of $g(\cdot)$ at $x$ if $g(\cdot)$ is only Lipschitz continuous. $D=\{l_1,l_2,\cdots,l_n\}$ is a dataset of $n$ samples.

%Vectors are in column form. For a vector $v$, we use $\|v\|_1$ to denote its $\ell_1$-norm and $\|v\|_2$ to denote its $\ell_2$-norm.
%It refers to the maximum impact of deleting an element in the dataset on the result of the query function $q(\cdot)$. Gaussian mechanism preservers $(\epsilon,\delta)$-differential privacy.

\subsection{Differential Privacy}
\cite{dwork2006calibrating} introduced the formal notion of differential privacy as follows.

\begin{definition}[Differential privacy]
A randomized mechanism $\mathcal{A}\!:\!\mathbb{D}^{n}\!\to\! \mathbb{R}$ is $(\epsilon,\delta)$-differential privacy ($(\epsilon,\delta)$-DP) if for all neighboring datasets $D,D'$ differing by one element and for all events $S$ in the output space of $\mathcal{A}$, we have
\begin{align*}
\textup{Pr}[\mathcal{A}(D)\in S]\leq e^{\epsilon}\,\textup{Pr}[\mathcal{A}(D')\in S]+\delta.
\end{align*}
And when $\delta=0$, $\mathcal{A}$ is $\epsilon$-differential privacy ($\epsilon$-DP).
\end{definition}

\begin{definition}[$\ell_2$-sensitivity \cite{Dwork:dp}]
For a function $q\!:\!\mathbb{D}^{n}\!\to\! \mathbb{R}$, the $\ell_2$-sensitivity $\bigtriangleup(q)$ of $q(\cdot)$ is defined as follows:
\begin{equation}
\bigtriangleup(q) = \max_{D,D'}\|q(D)-q(D')\|
\end{equation}
where $D,D'\in\mathbb{D}^{n}$ are a pair of neighboring datasets, which differ in a single entry.
\end{definition}

Sensitivity is a key indicator that determines the size of the noise added to the algorithm. That is, to achieve $(\epsilon,\delta)$-DP for a function $q\!:\!\mathbb{D}^{n}\!\to\! \mathbb{R}$, we usually use the following Gaussian mechanism, where the added noise is sampled from Gaussian distribution with variance that is proportional to the $\ell_2$-sensitivity of $q(\cdot)$.

\begin{definition}[Gaussian mechanism \cite{Dwork:dp}]
Given a function $q\!:\!\mathbb{D}^{n}\!\to\! \mathbb{R}$, the Gaussian mechanism $\mathcal{A}$ is defined as follows:
\begin{align*}
\mathcal{A}(D,q,\epsilon)=q(D)+v
\end{align*}
where $v$ is drawn from Gaussian distribution $N(0,\sigma^2I)$ with $\sigma\geq\frac{\sqrt{2\log(1.25/\delta)}\bigtriangleup(q)}{\epsilon}$. Here $\bigtriangleup(q)$ is the $\ell_2$-sensitivity of the function q, i.e., $\bigtriangleup(q)=\sup_{D\sim D'}\!\|q(D)-q(D')\|$.
\end{definition}

\subsection{R\'enyi Differential Privacy}
Although the definition of $(\epsilon,\delta)$-DP is widely used in the objective and output perturbation methods, the notion of R\'enyi differential privacy (RDP) \cite{mironov2017renyi} is more suitable for gradient perturbation methods including the proposed algorithms.

\begin{definition}[R\'enyi divergence \cite{mironov2017renyi}]
For two probability distributions $P$ and $Q$ defined on $\mathbb{R}$, the R\'enyi divergence of order $\alpha>1$ is defined as follows:
\[D_{\alpha}(P\parallel Q)\triangleq\frac{1}{\alpha-1}\log \textup{E}_{x\sim Q}\big(\frac{P(x)}{Q(x)}\big)^{\alpha}.\]
\end{definition}

%and for all events $S$ in the output space of $\mathcal{A}$
\begin{definition}[R\'enyi Differential Privacy (RDP) \cite{mironov2017renyi}]
A randomized mechanism $\mathcal{A}$ is $(\alpha,\beta)$-R\'enyi differentially private ($(\alpha,\beta)$-RDP) if for all neighboring datasets $D,D'$, we have
\begin{align*}
D_{\alpha}(\mathcal{A}(D)\|\mathcal{A}(D'))\leq\beta.
\end{align*}
That is, the R\'enyi divergence of the output of the function $\mathcal{A}$ is less than $\beta$.
\end{definition}

\begin{definition}[R\'enyi Gaussian Mechanism \cite{mironov2017renyi}]
Given a function $q:\mathbb{D}^{n}\!\to\! \mathbb{R}$, the Gaussian mechanism $\mathcal{A}=q(D)+v$ satisfies $(\alpha,\alpha\bigtriangleup^2\!(q)/(2\sigma^2))$-RDP, where $v\sim N(0,\sigma^2I)$.
\end{definition}

%We give the following relationship between RDP and $(\epsilon, \delta)$-DP \cite{mironov2017renyi}:
%\begin{proposition}
%[From RDP to $(\epsilon, \delta)$-DP] If a randomized mechanism $\mathcal{A}:\mathbb{D}^{n}\!\to\! \mathbb{R}$ satisfies $(\alpha,\beta)$-RDP, for any $\delta\in(0,1)$, $\mathcal{A}$ satisfies $(\beta+\log(1/\delta)/(\alpha-1),\delta)$-DP.
%\end{proposition}
%
%RDP has the following composition properties \cite{mironov2017renyi}:
%\begin{lemma} [RDP composition]
%If $k$ randomized mechanisms $\mathcal{A}_i:\mathbb{D}^{n}\!\to\! \mathbb{R}$ satisfy $(\alpha,\beta_i)$-RDP, then their composition $(\mathcal{A}_1(D),\dots,\mathcal{A}_k(D))$ satisfies $(\alpha,\sum_{i=1}^{k}\beta_i)$-RDP. Moreover, the input of the $i$-th mechanism can be based on the output of the first (i-1) mechanisms.
%\end{lemma}

\subsection{Nesterov's Accelerated Method}

\begin{algorithm}[t]
\caption{Nesterov's Accelerated Gradient Descent}
\label{alg_gp}
\renewcommand{\algorithmicrequire}{\textbf{Input:}}
\renewcommand{\algorithmicensure}{\textbf{Initialize:}}
\renewcommand{\algorithmicoutput}{\textbf{Output:}}
\begin{algorithmic}[1]
\ENSURE $\theta_1=1$, $x_0=y_1\in \mathbb{R}^d$, learning rate $\eta<1/L_F$, where $L_F$ is the Lipschitz constant for $\nabla F$.\\		
\FOR{$t = 1,2,...,T$}		
\STATE {$x_t = y_t - \eta\nabla F(y_t)$;}\\		
\STATE {$\theta_{t+1} = (1 + \sqrt{4\theta_t^2+1})/2$;}\\ 		
\STATE {$y_{t+1} = x_t + (\theta_{t}-1)(x_t-x_{t-1})/\theta_{t+1}$;}
\ENDFOR
\OUTPUT {$y_{T}$.}
\end{algorithmic}
\end{algorithm}

In \cite{nesterov:fast}, Nesterov presented a first-order minimization scheme with a global convergence rate $O(1/T^2)$ for solving Problem (\ref{eq:finite_sum_problem}). This convergence rate is provably optimal for the class of Lipschitz differentiable functionals. As shown in Algorithm~\ref{alg_gp}, the Nesterov method accelerates the gradient descent by using an overrelaxation step as follows:
\[y_{t+1} = x_t + \frac{\theta_t-1}{\theta_{t+1}}(x_t-x_{t-1})\]
where $\theta_{t+1} = (1 + \sqrt{4\theta_t^2+1})/2$ with the initial value $\theta_1\!=\!1$.

In 2014, Goldstein et al.\ \cite{goldstein2014fast} proposed an accelerated ADMM algorithm (AccADMM) by introducing Nesterov acceleration. They also proved that their algorithm has a global convergence rate of $O(1/T^2)$.

\section{Differentially Private ADMMs}
\label{sec4}
In this section, we propose two new deterministic differentially private ADMM algorithms for many machine learning problems such as the $\ell_1$-norm regularized and graph-guided fused Lasso. The proposed algorithms protect privacy by adding gradient perturbations. In particular, the second algorithm (DP-AccADMM) introduces the Nesterov acceleration into the first algorithm (DP-ADMM). Moreover, we also provide the privacy guarantees for the proposed algorithms.

\subsection{Differentially Private ADMM}
For solving the equality constrained minimization problem \eqref{equ002}, and the specific algorithmic steps of our deterministic differentially private ADMM (DP-ADMM) algorithm are presented in Algorithm~\ref{alg:1}.

Analogous to the general ADMM algorithm, our DP-ADMM algorithm updates and iterates the $x, y, u$ variables in an alternating fashion. But when updating $x_t$, we use the first-order approximation of $f(x)$ at $x_{t-1}$ with Gaussian noise (i.e., $f(x_{t-1})+\langle\nabla\! f(x_{t-1})+P_t,\,x\rangle$) to replace $f(x)$, where $P_t\!\sim\! N(0,\sigma^2I_{d_1})$ is the added Gaussian noise, and $\sigma$ is a noisy variance computed by privacy guarantee, which satisfies the Gaussian mechanism. Then we add a squared norm term $\frac{\|x-x_{t-1}\|_G^2}{2\eta}$ into the following proximal update rule of $x_{t}$,
\begin{equation}\label{equ007}
\begin{split}
&x_t=\arg\min_x \big\{\langle\nabla\! f(x_{t-1})+P_t,\,x\rangle+\frac{\rho}{2}\| Ax+By_t\\
&\qquad\qquad\quad\qquad\quad\;\;\;-c+u_{t-1}\|^2+\frac{\| x-x_{t-1}\|_G^2}{2\eta}\big\}
\end{split}
\end{equation}
where $\eta$ is a step-size or learning rate, $G=\gamma I-\eta\rho A^T\!A$, $\gamma\geq\gamma_{\min}\equiv\eta\rho\|A^T\!A\|_2+1$ to ensure that $G\succeq I$, and $\|z\|^{2}_{G}\!=\!z^{T}Gz$ with a given positive semi-definite matrix $G$ as in~\cite{ouyang2013stochastic,zheng2016fast}. Introducing this squared norm term can make the distance between adjacent iterates (i.e., $x_{t-1}$ and $x_t$) not be too far, and prevent the noise from affecting the iterates too much.

By using the linearized proximal point method~\cite{ouyang2013stochastic}, we can obtain the closed-form solution of $x_t$ in Eq.\ (\ref{equ007}) as follows:
\begin{align*}
x_t\!=\!x_{t-\!\!\:1}\!-\!\frac{\eta}{\gamma}[\nabla\! f(x_{t-\!\!\:1})\!+\!P_t\!+\!\rho A^T\!(Ax_{t-\!\!\:1}\!+\!By_t\!-\!c\!+\!u_{t-\!\!\:1})].
\end{align*}

As shown in Algorithm~\ref{alg:1}, the update rules of $y_{t}$ in Eq.\ (\ref{equ003}) and $u_{t}$ in Eq.\ (\ref{equ005}) remain unchanged for DP-ADMM. Here, $g(y)$ is the regularization term used in many machine learning problems, e.g., the sparsity inducing term $g(y)\!=\!\lambda\|Ay\|_1$. For instance, if it is the $\ell_1$-norm regularized term, then the closed-form solution can be easily obtained using the soft-thresholding operator~\cite{Donoho:soft}, and when it is the $\ell_2$-norm term, the closed-form solution can be obtained by derivation.

\begin{algorithm}[t]
\caption{DP-ADMM($f,g,x_0,T,\eta,\sigma$)}
\label{alg:1}
\renewcommand{\algorithmicrequire}{\textbf{Input:}}
\renewcommand{\algorithmicensure}{\textbf{Initialize:}}
\renewcommand{\algorithmicoutput}{\textbf{Output:}}
\begin{algorithmic}[1]
\REQUIRE {$f$ is $L_f$-smooth, learning rate $\eta$, and $T$.}\\
\ENSURE {$x_0$, $u_0=-\frac{1}{\rho}(A^T)^{\dagger}\nabla f(x_0)$;}\\
\FOR{$t=1,2,\ldots,T$}
\STATE {$y_t=\arg\min_y \big\{g(y)+\frac{\rho}{2}\| Ax_{t-1}+By-c+u_{t-1}\|^2\big\}$;}\\
\STATE {$x_t=\arg\min_x \big\{\langle\nabla\! f(x_{t-1})+P_t,\,x\rangle+\frac{\rho}{2}\| Ax+By_t$\\$\;\;-\,c+u_{t-1}\|^2+\frac{\| x-x_{t-1}\|_G^2}{2\eta}\big\}$, where $P_t\sim N(0,\sigma^2I_{d1})$;}\\
\STATE {$u_t=u_{t-1}+Ax_t+By_t-c$;}\\
\ENDFOR
%\STATE $x_*=x_T , y_*=y_T , u_*=u_T$;\\ ,y_0
\OUTPUT {$x_T,y_T$}\\
\end{algorithmic}
\end{algorithm}

\subsection{Privacy Guarantee Analysis}
In this subsection, we theoretically analyze both privacy guarantee and utility guarantee of DP-ADMM. To facilitate our discussion, we first make the following basic assumptions.

\begin{assumption}
For a convex and Lipschitz-smooth function $f$, there exists a constant $L_f$ such that $\|\nabla\! f(x_1)-\nabla\! f(x_2)\|\leq L_f\|x_1-x_2\|$ for any $x_1,x_2$.
\end{assumption}

\begin{assumption}
The matrix $A$ has full row rank. It makes sure that the matrix $A$ has a pseudo-inverse.
\end{assumption}

Through the Gaussian mechanism of RDP, we can get the relationship between $\alpha$ and $\beta$ of RDP and the variance $\sigma$ of the added noise. Then, we can obtain the relationship between $\epsilon,\delta$ and $\sigma$ by the conversion relationship between RDP and $(\epsilon,\delta)$-DP. Thus, we can get the size of Gaussian noise variance.

\begin{theorem}[Privacy guarantee]
\label{theo1}
For DP-ADMM, it satisfies $(\epsilon,\delta)$-differential privacy with some constants $c,\delta>0$ and $\mu\in(0,1)$, if
\begin{equation}\label{equ018}
\sigma^2=\frac{c^2\alpha T}{2n^2\epsilon\mu}
\end{equation}
where $\alpha=\frac{\log(1/\delta)}{(1-\mu)\epsilon}+1$.
\end{theorem}
The detailed proof of Theorem~\ref{theo1} is given in Appendix A.

Let ($x_*,y_*$) be an optimal solution of Problem (\ref{equ002}). Different from distributed differential privacy ADMMs \cite{huang2019dp,wang:admm}, by constructing the following convergence criterion $R(\tilde{x},\tilde{y})$,
\begin{align*}
R(\tilde{x},\tilde{y})&=f(\tilde{x})-f(x_*)-\big\langle\nabla\! f(x_*), \,\tilde{x}-x_*\big\rangle\\&\quad+g(\tilde{y})-g(y_*)-\big\langle g'(y_*),\,\tilde{y}-y_*\big\rangle
\end{align*}
where $\tilde{x}\!=\!\frac{1}{T}\sum_{t=1}^{T}\!x_t$ and $\tilde{y}\!=\!\frac{1}{T}\sum_{t=1}^{T}\!y_t$, we can complete the convergence analysis of DP-ADMM. Then by introducing the noise variance term and finding the number of iterations $T$ required to reach the stopping criterion, we can get the utility bound and gradient complexity of DP-ADMM. The utility bound is an important indicator to measure the performance of differential privacy algorithms including DP-ADMM. It usually reflects the minimum value that the algorithm can converge to. The lower the utility bound is, the smaller the loss can decrease and the better the trained model is.

Gradient complexity reflects the number of gradient calculations required by the model to reach the utility bound. It is usually proportional to the running time of the algorithm and has a certain relationship with the convergence rate of the algorithm. The smaller the gradient complexity is, the fewer the number of gradients need to be calculated.

\begin{theorem}[Utility guarantee]
\label{theo2}
In DP-ADMM, let $\sigma$ be defined in Eq.\ (\ref{equ018}), $\tilde{x}\!=\!\frac{1}{T}\sum_{t=1}^{T}x_t $, and $\tilde{y}\!=\!\frac{1}{T}\sum_{t=1}^{T}y_t$. Then, if the number of iterations $T=O(\frac{n\sqrt{\epsilon\mu}}{\sqrt{\alpha\eta d_1}})$, we can get the utility bound of our DP-ADMM algorithm as follows:
\begin{equation}
R(\tilde{x},\tilde{y})\leq O(\frac{\sqrt{\alpha\eta d_1}}{n\sqrt{\epsilon\mu}}).
\end{equation}
Moreover, the gradient complexity of our DP-ADMM algorithm is $O(\frac{n^2\sqrt{\epsilon\mu}}{\sqrt{\alpha\eta d_1}})$.
\end{theorem}

The detailed proof of Theorem~\ref{theo2} can be found in Appendix B. Theorem 2 shows that the privacy guarantee and utility guarantee of DP-ADMM. That is, the size of the added noise variance required by the algorithm to meet differential privacy requirement is given, and the minimum value of the algorithm's loss is obtained. Compared with the theoretical analysis of traditional ADMMs, we find that the convergence rate of DP-ADMM does not change. This means that differential privacy protection does not affect the convergence rate of the optimization algorithm or the speed at which the optimization algorithm converges to the optimal value, but the optimal value that the algorithm converges to.

\subsection{Differentially Private Accelerated ADMM}
\label{sec321}
Non-accelerated ADMMs including DP-ADMM can be used to solve many machine learning problems with equality constraints, but it has a flaw that the convergence speed is not fast enough as in the non-private case. Especially when approaching to the exact solution, the convergence speed is very slow. Therefore, we propose an accelerated version of our DP-ADMM algorithm, called DP-AccADMM.

\begin{algorithm}[t]
\caption{DP-AccADMM($f,g,\hat{x}_0,T,\eta,\sigma,\alpha_0$)}
\label{alg:3}
\renewcommand{\algorithmicrequire}{\textbf{Input:}}
\renewcommand{\algorithmicensure}{\textbf{Initialize:}}
\renewcommand{\algorithmicoutput}{\textbf{Output:}}
\begin{algorithmic}[1]
\REQUIRE {$f$ is $L_f$-smooth, learning rate $\eta$, and $T$.}
\ENSURE {$\hat{x}_0\!=\!x_0$, $u_0\!=\!\hat{u}_0\!=\!-\frac{1}{\rho}(A^T)^{\dagger}\nabla\! f(\hat{x}_0)$, and $\theta_1\!=\!1$.}
\FOR {$t=1,2,\ldots,T$}
\STATE {$y_t=\arg\min_y \big\{g(y)+\frac{\rho}{2}\| A\hat{x}_{t-1}+By-c+\hat{u}_{t-1}\|^2\big\}$;}
\STATE {$x_t=\arg\min_x \big\{\langle\nabla\! f(\hat{x}_{t-1})+P_t,\,x\rangle+\frac{\rho}{2}\| Ax+By_t-c+\hat{u}_{t-1}\|^2+\frac{\| x-\hat{x}_{t-1}\|_G^2}{2\eta}\big\}$, where $P_t\sim N(0,\sigma^2I_{d1})$;}
\STATE {$u_t=\hat{u}_{t-1}+Ax_t+By_t-c$;}
\STATE {$\theta_{t+1}=(1+\sqrt{1+4\theta_{t}^2})/{2}$;}
\STATE {$\hat{x_t}=x_{t}+\frac{\theta_{t}-1}{\theta_{t+1}}(x_t-x_{t-1})$;}
\STATE {$\hat{u_t}=u_{t}+\frac{\theta_{t}-1}{\theta_{t+1}}(u_t-u_{t-1})$;}
\ENDFOR
\OUTPUT {$x_T,y_T$}
\end{algorithmic}
\end{algorithm}

Inspired by the accelerated ADMM algorithm \cite{goldstein2014fast} using the Nesterov method, we introduce the Nesterov accelerated technique into our DP-ADMM algorithm. The detailed steps of our accelerated differentially private ADMM (DP-AccADMM) algorithm are presented in Algorithm~\ref{alg:3}. In particular, we add a weight coefficient during the iteration process to linearly combine the two adjacent iterates, so that the current iterate can be updated by using the information of the previous iterate. That is, the current descent direction is the combination of the current optimal direction and the previous descent direction, which will make the current iterate to move forward a distance on the previous descent direction. Note that the weight coefficient for momentum acceleration changes with the increase of iterations. In our DP-AccADMM algorithm, we set the weight coefficient $\theta_t$ at the $t$-th iteration as follows:
$$\theta_{t+1}=\frac{1+\sqrt{1+4\theta_{t}^2}}{2}$$ where $\theta_1\!=\!1$. And the momentum accelerated update rule of  $\hat{x}_t$ is defined as:
\begin{equation}
\hat{x}_t=x_t+\frac{\theta_{t}-1}{\theta_{t+1}}(x_t\!-\!x_{t-1}).
\end{equation}
Such an accelerated scheme is consistent with the acceleration method proposed by Nesterov \cite{nesterov:co} and the accelerated ADMM algorithm proposed by Goldstein et al.\ \cite{goldstein2014fast}. In addition, the dual variable $\hat{u_t}$ also has the same accelerated scheme as $\hat{x}_t$, as shown in Step 7 of Algorithm~\ref{alg:3}.

Although we do not give the theoretical analysis of DP-AccADMM, we conduct many experiments to compare the performance of the two algorithms, DP-AccADMM and DP-ADMM, whose noise level is the same, as shown in Theorem~\ref{theo1}. We will report many experimental results and give some discussions in the next section.

\begin{figure*}[t]
\centering
\subfigure[a8a]{\includegraphics[width=0.489\columnwidth]{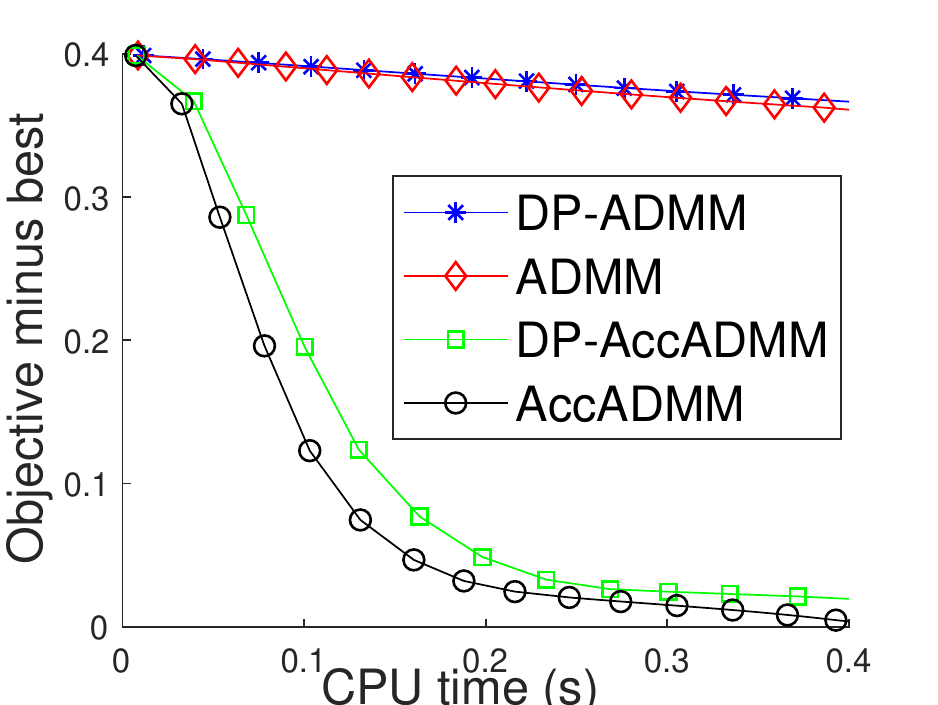}}\,
\subfigure[a9a]{\includegraphics[width=0.489\columnwidth]{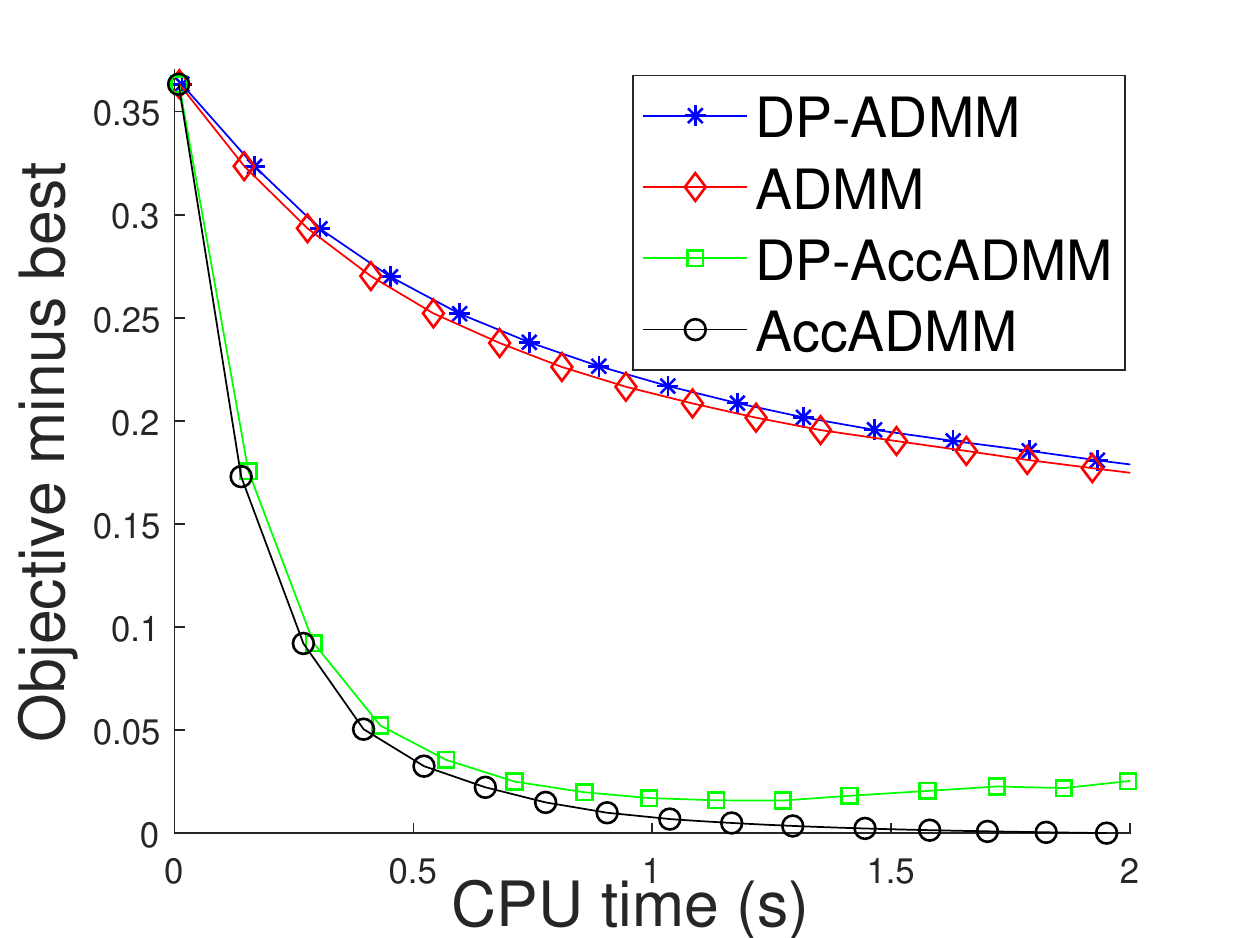}}\,
\subfigure[bio\_train]{\includegraphics[width=0.489\columnwidth]{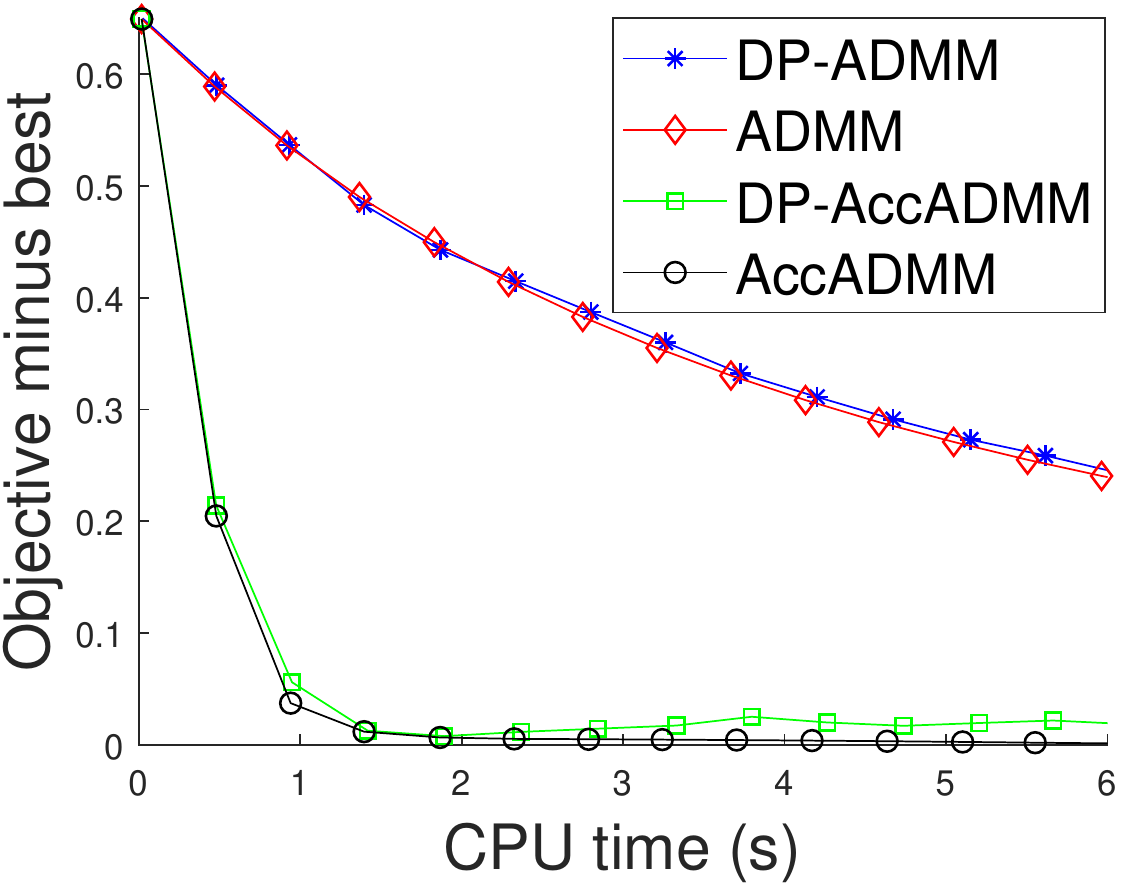}}\,
\subfigure[ijcnn1]{\includegraphics[width=0.489\columnwidth]{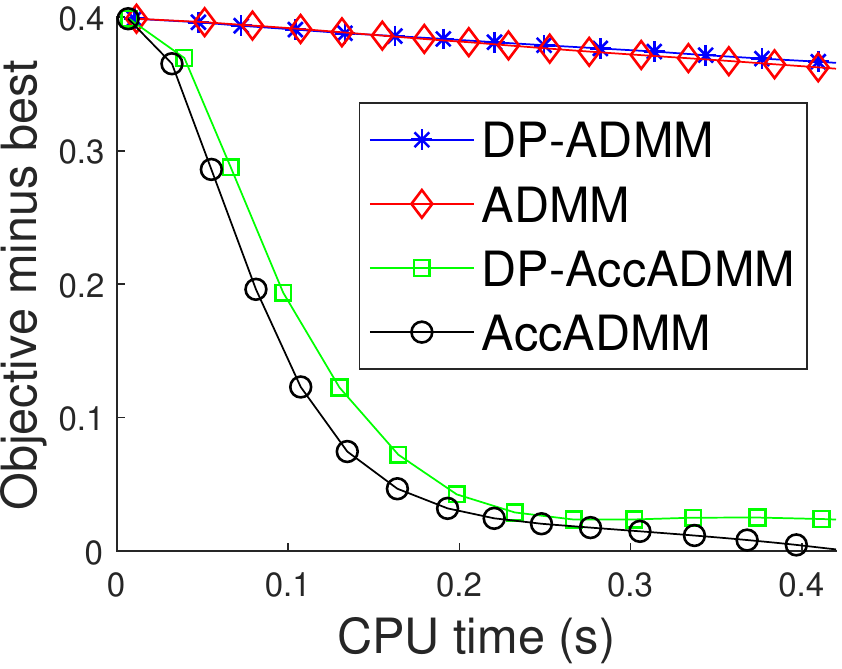}}
\vspace{-3mm}

\caption{Comparison of the proposed differential privacy ADMM algorithms (including DP-ADMM and DP-AccADMM) and their non-differentially private counterparts for solving graph-guided fused Lasso problems on the four data sets. The $x$-axis is the objective value minus the minimum value, and the $y$-axis denotes the running time (seconds).}
\label{figure_lr}
\end{figure*}

\section{Experimental Results}
\label{sec5}
In this section, we evaluate the performance of our DP-ADMM and DP-AccADMM algorithms and report some experimental results on four publicly available datasets, as shown in Table \ref{tab01}. We use our two algorithms to solve the general convex graph-guided fused Lasso problem, and compare our algorithms with their counterparts: ADMM and Acc-ADMM. We first show the convergence speed performance of all the methods in terms of CPU time (seconds) during the training process, and then show the performance of model training and testing as the privacy parameter $\epsilon$ changes.

\subsection{Graph-Guided Fused Lasso}
To evaluate performance of the proposed algorithms, we consider the following $\ell_1$-norm regularized graph-guided fused Lasso problem,
\begin{equation} \label{the graph-guided fuzed Lasso problem}
\min_x \Big\{\frac{1}{n}\sum_{i=1}^{n} f_i(x,l_i)+\lambda\|y\|_1 ,\;\textup{s.t.,}\; Ax-y = 0\Big\}
\end{equation}
where $f_i$ is the logistic loss function on the feature-label pair $(l_i,m_i)$, i.e., $\log(1+\exp(-m_il_i^Tx))$, and $\lambda$ is the regularization parameter. We set $A=[W;I]$ as in \cite{zheng2016fast,Ouyang2012Stochastic,azadi2014towards}, where $W$ is the sparsity pattern
of the graph obtained by sparse inverse covariance selection \cite{El2008Model}. We used four publicly available datasets in our experiments, as listed in Table \ref{tab01}.

\begin{table}[!t]
	\caption{Summary of data sets and regularization parameters used in our experiments.}
     \label{tab01}
	\centering
	\setlength{\tabcolsep}{8.6pt}
	\begin{tabular}{ccccc}
		\toprule  		
		Data sets & {$\#$training} & {$\#$test}  & {$\#$features} & {$\lambda_1$} \\
		\hline		
		a9a  & 32,561 & 16,281  & 123 & 1e-5   \\
		a8a  & 22,696 & 9,865  & 123 & 1e-5   \\
		ijcnn1  & 49,990 & 91,701  & 22 & 1e-5   \\
		bio\_train  & 72,876 & 72,875  & 74 & 1e-5   \\		
		\bottomrule
	\end{tabular}	
\end{table}

\subsection{Parameter Setting}
We fix $\delta\!=\!10^{-3}$ for all the experiments, and we choose $\mu=0.5$ and $\rho=1$, where $\mu$ is an intermediate constant in the solution process and $\rho$ is the penalty parameter of the ADMM algorithms. But in our theoretical analysis, $\rho$ will not influence the utility bound of our differentially private algorithms. Thus, we fix $\rho$ as a constant. Moreover, we set $\gamma=1$ and set $\eta$ as a constant on each dataset but differs among these datasets.

%Actually, $\rho$ is a variate.

\subsection{Objective Value Decreasing with CPU Time}

\begin{figure*}[!ht]
\centering
\includegraphics[width=0.485\columnwidth]{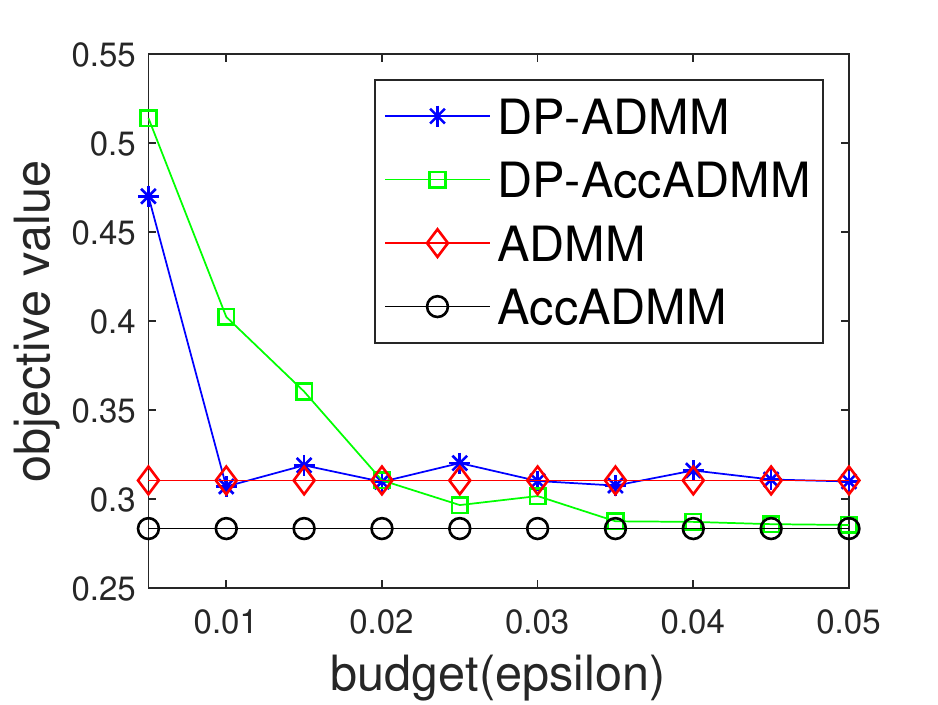}\,
\includegraphics[width=0.485\columnwidth]{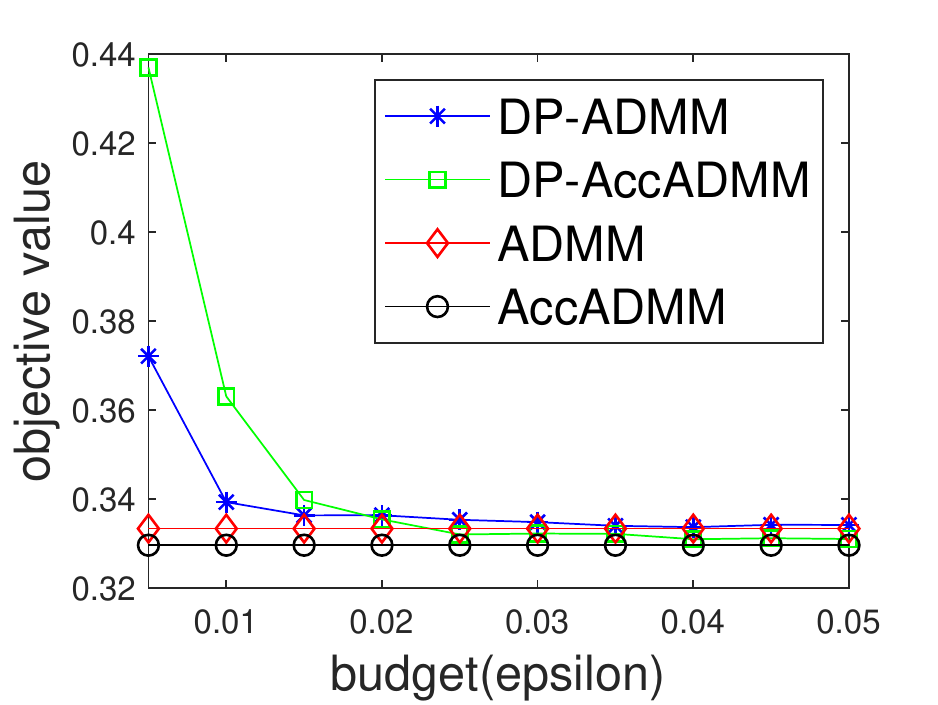}\,
\includegraphics[width=0.485\columnwidth]{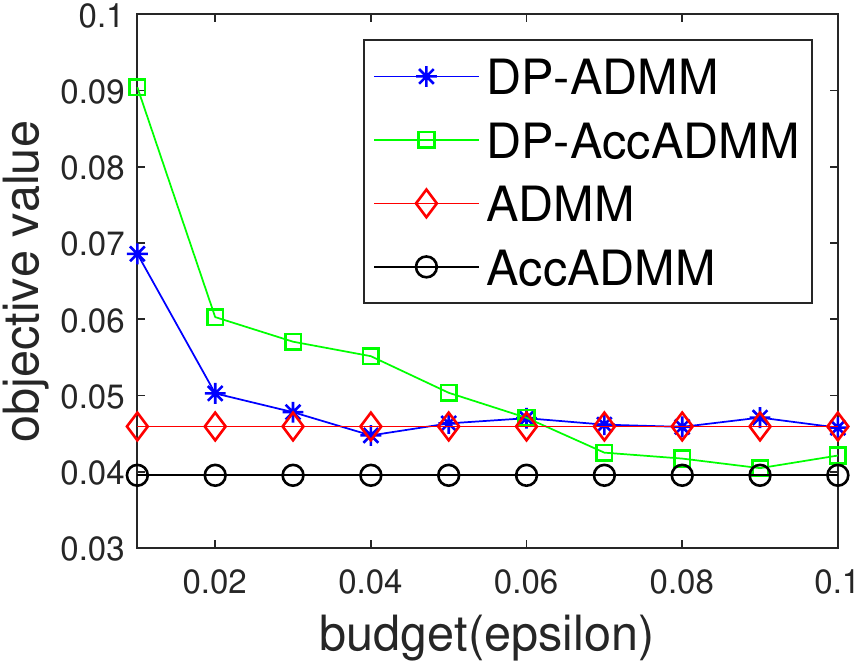} \,
\includegraphics[width=0.485\columnwidth]{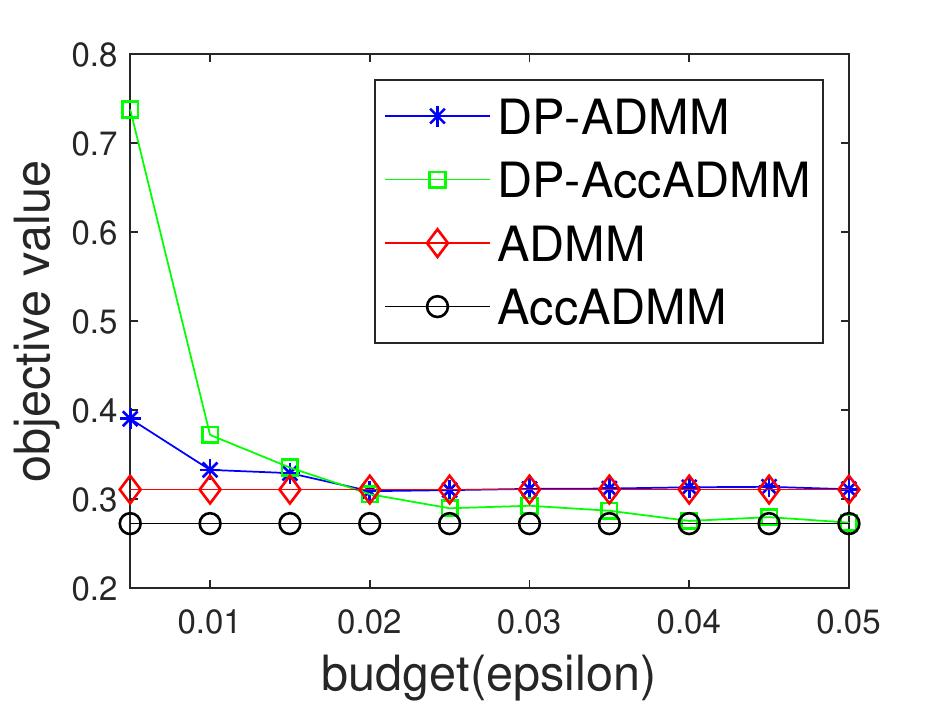}
\subfigure[a8a]{\includegraphics[width=0.485\columnwidth]{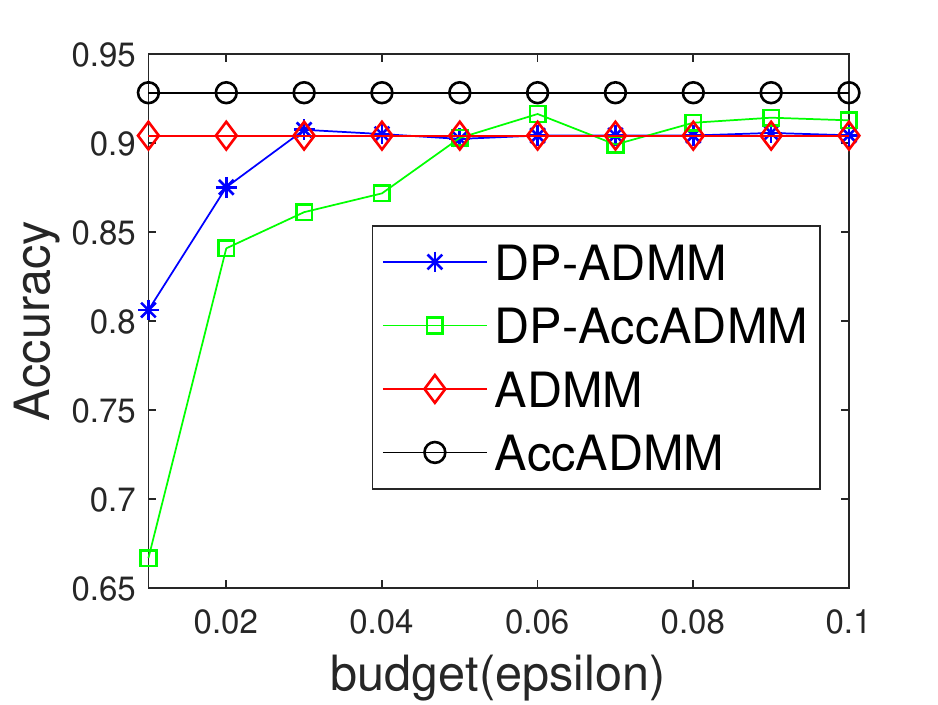}}\,
\subfigure[a9a]{\includegraphics[width=0.485\columnwidth]{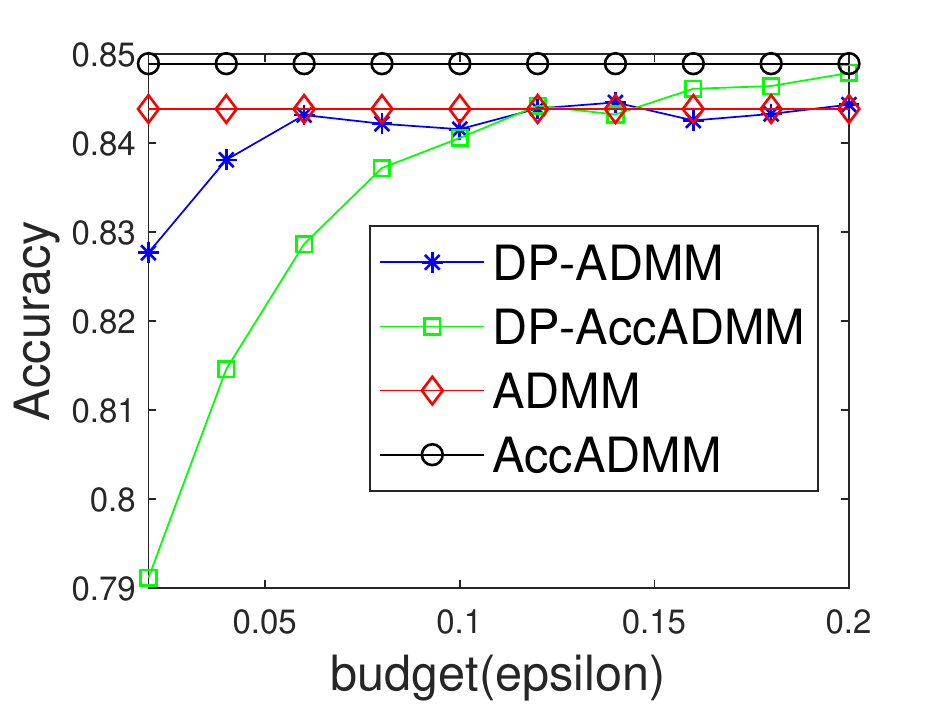}}\,
\subfigure[bio\_train]{\includegraphics[width=0.485\columnwidth]{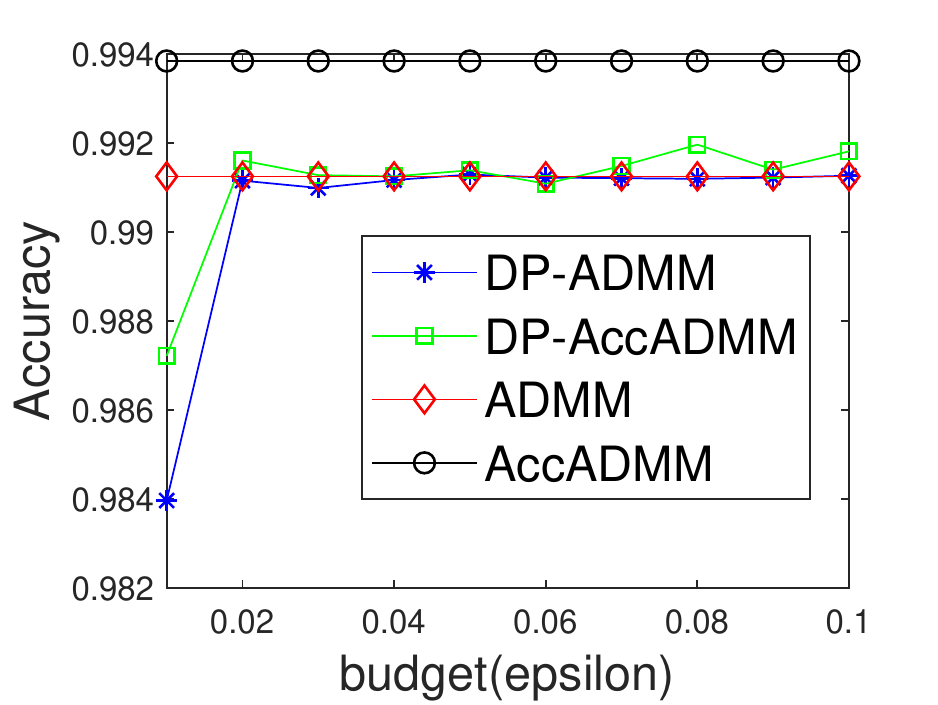}}\,
\subfigure[ijcnn1]{\includegraphics[width=0.485\columnwidth]{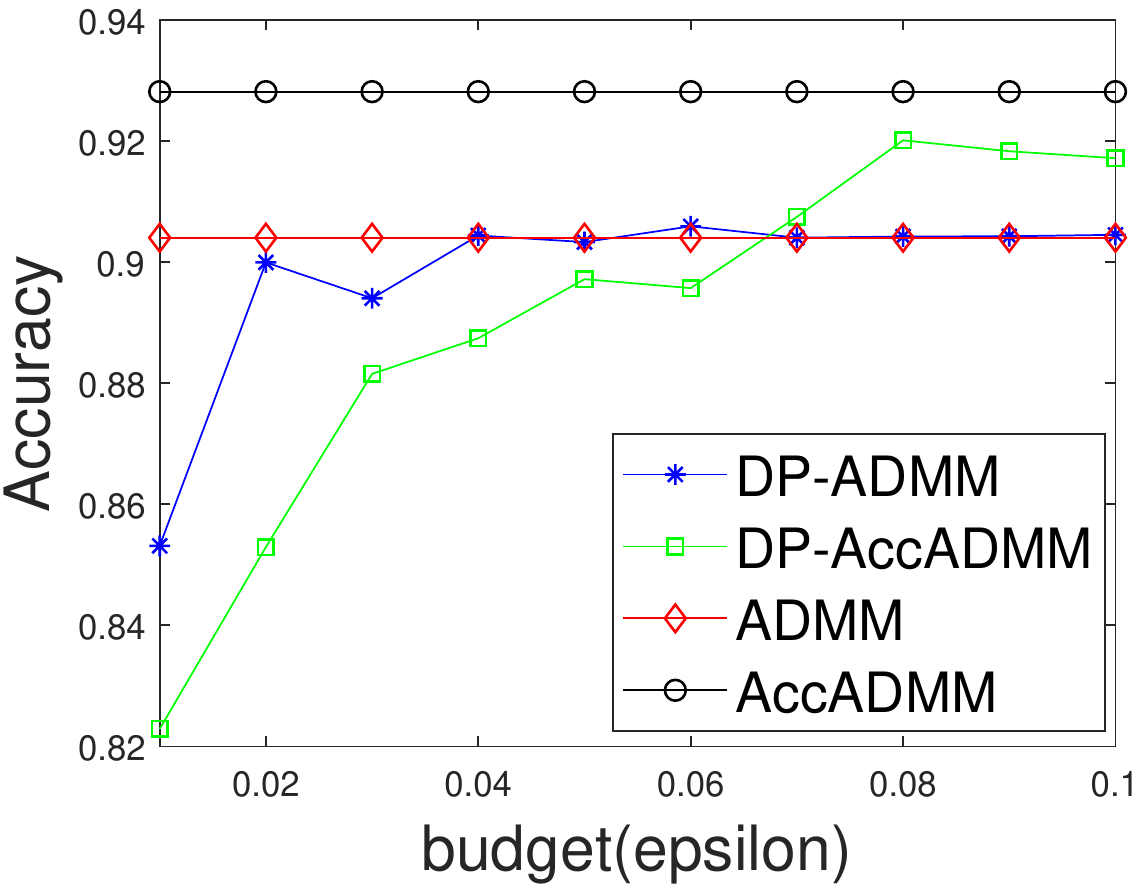}}
\vspace{-3mm}

\caption{Graph-guided fused Lasso results of all the algorithms with different privacy budgets, $\epsilon$. Top: Objective value vs budget; Bottom: Classification accuracy vs budget.}
\label{figure_epsilon}
\end{figure*}

Fig.\ \ref{figure_lr} plots the objective gap (i.e., the objective value minus the minimum value) of the differential privacy ADMM algorithms as the number of iterations increases. Actually, the accelerated ADMM algorithm can achieve the convergence rate of $O(1/T^2)$, and it is much faster than the traditional ADMM algorithm. In our experiments, it can be seen that the convergence speed of DP-AccADMM is almost the same as that of its non-differential privacy counterpart, AccADMM. In addition, we can see that there are two gaps between two pairs of algorithms. These gaps are the embodiment of the utility boundary in the utility analysis in the experiments. And the gap between DP-AccADMM and AccADMM is larger than that between DP-ADMM and ADMM. It means that the Nesterov acceleration may increase the utility bound of the algorithm, which will reduce the utility of the algorithm. Therefore, we do another experiment to show the performance of test accuracy of all these algorithms.

\subsection{Performance on Simulated Data}
Fig.\ \ref{figure_epsilon} plots the objective value and test accuracy of the differential privacy ADMM algorithms with different privacy budgets $\epsilon$ for solving $\ell_1$-norm regularized graph-guided fused Lasso. We can see that AccADMM converges much faster than ADMM in terms of objective value. DP-AccADMM performs worse than DP-ADMM, when the private budget $\epsilon$ is small (e.g., $\epsilon\!=\!0.01$). But with the increase of the budget $\epsilon$ (e.g., $\epsilon\!=\!0.07$ on bio\_train), DP-AccADMM outperforms DP-ADMM in terms of both objective value and test accuracy. As the private budget increases, the objective value of DP-AccADMM gradually approaches to that of AccADMM, and the objective value of DP-ADMM approaches to that of ADMM. Thus, the Nesterov acceleration usually leads to a worse utility, when $\epsilon$ is small. However, when $\epsilon$ becomes larger, DP-AccADMM gets better and outperforms DP-ADMM.

From test accuracy results, we can see that DP-AccADMM also performs worse than DP-ADMM, when the private budget $\epsilon$ is small. With the increase of $\epsilon$, the test accuracy of DP-AccADMM gradually improves, approaching or even exceeding the test accuracy of DP-ADMM. Moreover, the test accuracy of DP-AccADMM gradually approaches to that of AccADMM and the test accuracy of DP-ADMM gradually approaches to that of ADMM. Thus, for test accuracy, we can see that the Nesterov acceleration technique usually leads to a worse test accuracy, when the private budget $\epsilon$ is small (e.g., $\epsilon\!=\!0.02$). However, with the increase of $\epsilon$ (e.g., $\epsilon\!=\!0.08$ on a8a and ijcnn1), the test accuracy of DP-AccADMM gradually increases and outperforms DP-ADMM until it gradually approaches to the test accuracy of AccADMM.

\section{Conclusions and Further Work}
\label{sec6}
In this paper, we proposed two efficient differentially private ADMM algorithms with the guarantees of $(\epsilon,\delta)$-DP. The first algorithm, DP-ADMM, uses gradient perturbation to achieve $(\epsilon,\delta)$-DP. Moreover, we also provided its privacy analysis and utility analysis through Gaussian mechanism and convergence analysis. The second algorithm, DP-AccADMM, uses the Nesterov method to accelerate the proposed DP-ADMM algorithm. All the experimental results showed that DP-AccADMM has a much faster convergence speed. In particular, DP-AccADMM can achieve a higher classification accuracy, when the privacy budget $\epsilon$ reaches a certain threshold.

Similar to DP-SVRG~\cite{Wang2018Differentially}, we can extend our deterministic differentially private ADMM algorithms to the stochastic setting for solving large-scale optimization problems as in~\cite{liu:asadmm}. As for the theoretical analysis of DP-AccADMM, we will complete it as our future work. Moreover, we can employ some added noise variance decay schemes as in \cite{ding:admm} to reduce the negative effects of gradient perturbation, and provide stronger privacy guarantees and better utility~\cite{bellet2018}. In addition, an interesting direction of future work is to extend our differentially private algorithms and theoretical results from the two-block version to the multi-block ADMM case~\cite{chen:admm}.

\appendix
\subsection*{Appendix A: Proof of Theorem 1}
Before giving the proof of Theorem 1, we present the following lemmas. We first give the following relationship between RDP and $(\epsilon, \delta)$-DP \cite{mironov2017renyi}:

%We give the following relationship between RDP and $(\epsilon, \delta)$-DP \cite{mironov2017renyi}:
%\begin{proposition}
%[From RDP to $(\epsilon, \delta)$-DP] If a randomized mechanism $\mathcal{A}:\mathbb{D}^{n}\!\to\! \mathbb{R}$ satisfies $(\alpha,\beta)$-RDP, for any $\delta\in(0,1)$, $\mathcal{A}$ satisfies $(\beta+\log(1/\delta)/(\alpha-1),\delta)$-DP.
%\end{proposition}
%
%RDP has the following composition properties \cite{mironov2017renyi}:
%\begin{lemma} [RDP composition]
%If $k$ randomized mechanisms $\mathcal{A}_i:\mathbb{D}^{n}\!\to\! \mathbb{R}$ satisfy $(\alpha,\beta_i)$-RDP, then their composition $(\mathcal{A}_1(D),\dots,\mathcal{A}_k(D))$ satisfies $(\alpha,\sum_{i=1}^{k}\beta_i)$-RDP. Moreover, the input of the $i$-th mechanism can be based on the output of the first (i-1) mechanisms.
%\end{lemma}

\begin{lemma}[From RDP to $(\epsilon, \delta)$-DP]
If a random mechanism $\mathcal{A}:\mathbb{D}^{n}\!\to\! \mathbb{R}$ satisfies $(\alpha,\beta)$-RDP, then for any $\delta\in(0,1)$, $\mathcal{A}$ satisfies $(\beta+\log(1/\delta)/(\alpha-1),\delta)$-DP.
\end{lemma}

This lemma shows the conversion relationship between $(\alpha,\beta)$-RDP and $(\epsilon,\delta)$-DP. And RDP has the following composition properties \cite{mironov2017renyi}:

\begin{lemma}
For $i\in[k]$, if $k$ random mechanisms $\mathcal{A}_i:\mathbb{D}^{n}\!\to\! \mathbb{R}$ satisfy $(\alpha,\beta_i)$-RDP, then their composition $(\mathcal{A}_1(D),\dots,\mathcal{A}_k(D))$ satisfies $(\alpha,\sum_{i=1}^{k}\beta_i)$-RDP. Moreover, the input of the $i$-th mechanism can be based on the output of the first (i-1) mechanisms.
\end{lemma}

This lemma shows the changes of the coefficients $\alpha,\beta$ that satisfy the RDP during the compounding process of a series of random mechanisms of the algorithm iteration.

\begin{lemma}[\cite{mironov2017renyi}]
Given a function $q:\mathbb{D}^{n}\!\to\! \mathbb{R}$, the Gaussian Mechanism $\mathcal{A}=q(D)+v$ satisfies $(\alpha,\alpha\bigtriangleup^2(q)/(2\sigma^2))$-RDP, where $v\sim N(0,\sigma^2I)$.
\end{lemma}
This lemma shows the relationship between the satisfied RDP coefficients $\alpha,\beta$ and the size of added noise variance $\sigma$.

Proof of Theorem 1:
\begin{proof}
$x_T\sim(\alpha,\beta)$-RDP $\to$ $(\beta+\frac{\log(1/\delta)}{\alpha-1},\delta)$-DP $=$ $(\epsilon,\delta)$-DP, where
\begin{equation}
\beta+\frac{\log(1/\delta)}{\alpha-1}=\epsilon.
\end{equation}
Let $\beta=\mu\epsilon$, $\frac{\log(1/\delta)}{\alpha-1}=(1-\mu)\epsilon$, and $\mu\sim(0,1)$, then
\begin{equation}
\alpha=\frac{\log(1/\delta)}{(1-\mu)\epsilon}+1.
\end{equation}
We can obtain
\begin{center}
$x_T\sim(\frac{\log(1/\delta)}{(1-\mu)\epsilon}+1,\mu\epsilon)$-RDP.
\end{center}
According to Lemma 2, we have
\begin{center}
$\mathcal{A}_k\sim(\frac{\log(1/\delta)}{(1-\mu)\epsilon}+1,\frac{\mu\epsilon}{T})$-RDP, $\qquad x_k\sim(\frac{\log(1/\delta)}{(1-\mu)\epsilon}+1,\frac{k\mu\epsilon}{T})$-RDP.\\
\end{center}
According to Definition 2 (i.e., the $L_2$-sensitivity of the function $q$), let $\max_i\|\nabla\! f_i(x)\|=c$, then we have $\bigtriangleup(q)=\frac{c}{n}$.	And according to Lemma 3, we have
\begin{center}
$\mathcal{A}_k\sim(\alpha,\frac{c^2\alpha}{2n^2\sigma^2})$-RDP.
\end{center}
Therefore, we can obtain
\begin{equation}
\label{equ012}
\sigma^2=\frac{c^2\alpha T}{2n^2\epsilon\mu}.
\end{equation}
That is, when the size of the noise variance we added satisfies Eq.\ (\ref{equ012}), the proposed DP-ADMM algorithm satisfies $(\epsilon,\delta)$-DP. This completes the proof.
\end{proof}

\subsection*{Appendix B: Proof of Theorem 2}
Before giving the proof of Theorem 2, we first present the following lemmas.

\begin{lemma}[\cite{zheng2016fast}]
$u_*=-\frac{1}{\rho}(A^T)^\dagger\nabla\! f(x_*)$.
\end{lemma}
This lemma presents the relationship between the optimal solution of the dual problem and the optimal solution of the original problem. Let us simply express $x_t$ with $x_{t-1}$ and the gradient term, which is convenient for our subsequent proof.
\begin{equation}
\begin{split}
\!\!\!\!\!x_t=&\arg\min_x \,\Big\{\langle\nabla\! f(x_{t-1})+P_t,\,x\rangle\\&\quad+\frac{\rho}{2}\| Ax+By_t-c+u_{t-1}\|^2+\frac{\| x-x_{t-1}\|^2_G}{2\eta}\Big\}.\!\!
\end{split}
\end{equation}
Setting the derivative with respect to $x$ at $x_t$ to zero, we have
\begin{equation}
\begin{split}
&\nabla\! f(x_{t-1})+P_t+\rho A^T(Ax+By_t-c+u_{t-1})\\&\qquad\quad\qquad+\frac{1}{\eta}G(x-x_{t-1})=0.
\end{split}
\end{equation}
Let
\begin{align*}
v_t&=\nabla\! f(x_{t-1})+P_t,\; P_t\sim N(0,\sigma^2I_{d1}),\\
q_t&=\rho A^T(Ax+By_t-c+u_{t-1}),\\
g_t&=v_t+q_t.\\	
\end{align*}
Then $g_t+\frac{1}{\eta}G(x-x_{t-1})=0$, and $x_t=x_{t-1}-\eta G^{-1}g_t$. Thus, we get a simple representation of $x_t$.

\begin{lemma}[\cite{zheng2016fast}]
If $0\le\eta\le\frac{1}{L_f}$, then we have
\begin{align*}
f(x)+q_t^T(x-x_t)\le &f(x_t)+g_t^T(x-x_{t-1})+\frac{\eta}{2}\| g_t\|^2_{G^{-1}}\\&+(v_t-\nabla\! f(x_{t-1}))^T(x_t-x).
\end{align*}
\end{lemma}

\begin{lemma}[\cite{zheng2016fast}]
Let $\alpha_t=\rho(u_t-u_*)$, we have
\begin{align*}
&2\eta[g(y_t)\!-\!g(y_*)\!-\!g'(y_*)^T(y_t\!-\!y_*)\!+\!(B^T\!\alpha_t)^T(y_*\!-\!y_t)]\\ \!\!\leq\;&\eta\rho[\| Ax_{t-1}+By_*-c\|^2-\|Ax_{t-1}+By_*-c\|^2\\&+\| u_t-u_{t-1}\|^2].
\end{align*}
\end{lemma}

\begin{lemma}[\cite{zheng2016fast}]
\begin{align*}
&\;2\eta[-(Ax_t+By_t-c)^T\alpha_t]\\
=&\;\eta\rho[\|  u_{t-1}-u_*\| ^2-\|  u_t-u_*\| ^2-\|  u_t-u_{t-1}\| ^2].
\end{align*}
\end{lemma}

Proof of Theorem 2:
\begin{proof}
\begin{align*}
&\quad\ \|x_t-x_*\|^2_G\quad\\&=\|  x_{t-1}-x_*\| ^2_G-2\eta(x_{t-1}-x_*)^Tg_t+\eta^2\|g_t\|^2_{G^{-1}}\\
&\leq\|  x_{t-1}-x_*\|^2_G-2\eta[f(x_t)-f(x_*)]\\&\quad-2\eta[v_t-\nabla\! f(x_{t-1})]^T(x_t-x_*)+2\eta q_t^T(x_*-x_t).
\end{align*}
Below let us bound $-2\eta(v_t-\nabla f(x_{t-1}))^T(x_t-x_*)$. Let
\begin{align*}
\psi_t(x)&=\frac{\rho}{2}\| Ax+By_t-c+u_{t-1}\|^2+\frac{1}{2\eta}\| x-x_{t-1}\|^2_{G-I},\\
\bar{x}&=\textup{prox}_{\eta\psi_t}(x_{t-1}-\eta\nabla f(x_{t-1}))
\end{align*}
where $\textup{prox}_{\eta r}(y)=\min_x\big\{ r(x)+\frac{1}{2\eta}\| x-y\|^2\big\}$, then we have
\begin{align*}
x_t&=\arg\min_x \,\Big\{v_t^Tx+\frac{\rho}{2}\| A x+By_t-c+u_{t-1}\|^2\\&\quad\qquad\quad\qquad\qquad\qquad\qquad\;\;+\frac{\| x-x_{t-1}\|^2_G}{2\eta}\Big\}\\
&=\arg\min_x \,\Big\{\eta v_t^Tx\!+\!\frac{\eta\rho}{2}\| A x\!+\!By_t\!-\!c\!+\!u_{t-1}\|^2\!\\&\qquad\quad\qquad\quad+\!\frac{\| x\!-\!x_{t-1}\|^2_{G-I}}{2}\!+\!\frac{\| x\!-\!x_{t-1}\|^2}{2}\Big\}\\
&=\arg\min_x \,\Big\{\eta\psi_t(x)+\frac{1}{2}\| x-(x_{t-1}-\eta v_t)\|^2\Big\}.
\end{align*}
Therefore,
\begin{equation}
x_t=\textup{prox}_{\eta\psi_t}(x_{t-1}-\eta v_t).
\end{equation}
Then, we have
\begin{align*}
&\quad-2\eta(v_t-\nabla\! f(x_{t-1}))^T(x_t-x_*)\\
&=-2\eta(v_t\!-\!\nabla\! f(x_{t-1}))^T\!(x_t\!-\!\bar{x})\!-\!2\eta(v_t\!-\!\nabla\! f(x_{t-1}))^T\!(\bar{x}\!-\!x_*\!\!\:)\\
&\le2\eta^2\| v_t-\nabla\! f(x_{t-1})\|^2-2\eta(v_t-\nabla f(x_{t-1}))^T(\bar{x}-x_*\!\!\:)
\end{align*}
and
\begin{align*}
&\quad\;\,\| x_t-x_*\|^2_G-2\eta q_t^T(x_*-x_t)\\&\le\| x_{t-1}-x_*\|^2_G-2\eta(f(x_t)-f(x_*))\\&\quad+2\eta^2\| v_t-\nabla\! f(x_{t-1})\|^2-2\eta(v_t-\nabla\! f(x_{t-1}))^T(\bar{x}-x_*).
\end{align*}
Since $v_t\!-\!\nabla\! f(x_{t-1})\!=\!P_t$, then $ \mathbb{E}\| P_t\|^2\!=\!\sigma^2d_{1}$, and $\mathbb{E}(P_t)\!=\!0$. Taking expectations on both sides of the above inequalities, we have
\begin{align*}
&\quad\ \| x_t-x_*\|^2_G-2\eta q_t^T(x_*-x_t)\\&\le\| x_{t-1}-x_*\|^2_G-2\eta(f(x_t)-f(x_*))+2\eta^2\sigma^2d_{1},
\end{align*}
\begin{align*}
&\quad\ 2\eta[f(x_t)-f(x_*)-q_t^T(x_*-x_t)]\\&\le\| x_{t-1}-x_*\|^2_G-\| x_t-x_*\|^2_G+2\eta^2\sigma^2d_{1}.
\end{align*}
Since
\begin{align*}
&2\eta[f(x_t)-f(x_*)-q_t^T(x_*-x_t)]\\=\;&2\eta[f(x_t)\!-\!f(x_*)\!-\!\nabla\! f(x_*)^T\!(x_t\!-\!x_*)\!-\!(A^T\!\alpha)^T\!(x_*\!-\!x_t)]
\end{align*}
then
\begin{align*}
&2\eta[f(x_t)\!-\!f(x_*)\!-\!\nabla\! f(x_*)^T\!(x_t\!-\!x_*)\!-\!(A^T\!\alpha)^T\!(x_*\!-\!x_t)]\\ \leq &\;\| x_{t-1}-x_*\|^2_G-\| x_t-x_*\|^2_G+2\eta^2\sigma^2d_{1}.
\end{align*}
According to Lemma 7, we have
\begin{align*}
&2\eta[g(y_t)-g(y_*)-g'(y_*)^T(y_t-y_*)+\alpha_t^T(y_*-y_t)]\\ \leq &\;\eta\rho\big[\| Ax_{t-1}+By_*-c\|^2-\| Ax_t+By_*-c\|^2\\&+\| u_t-u_{t-1}\|^2\big].
\end{align*}
%Let
%\begin{align*}
%R(x,y)&=f(x)-f(x_*)-\nabla\! f(x_*)^T(x-x_*)\\&\quad+g(y)-g(y_*)-g'(y_*)^T(y-y_*).
%\end{align*}
And we have
\begin{align*}
0=&-(A^T\alpha_t)^T(x_*-x_t)-(B^T\alpha_t)^T(y_*-y_t)\\&-(Ax_t+By_t-c)^T\alpha_t.
\end{align*}
Thus, according to Lemma 8, we have
\begin{align*}
2\eta R(x_t,y_t)\le&\,\| x_{t-1}-x_*\|^2_G-\| x_t-x_*\|^2_G+2\eta^2\sigma^2d_{1}\\
&-2\eta(Ax_t\!-\!y_t)^T\!\alpha_t\!+\!\eta\rho[\| Ax_{t-1}\!+\!By_*\!-\!c\|^2\\
&-\| Ax_t+By_*-c\|^2+\| u_t-u_{t-1}\|^2]\\=&\;\| x_{t-1}-x_*\|^2_G-\| x_t-x_*\|^2_G+2\eta^2\sigma^2d_{1}\\
&+\!\eta\rho[\| Ax_{t-1}\!+\!By_*\!\!-\!c\|^2\!-\!\| Ax_t\!+\!By_*\!\!-\!c\|^2\!\\&+\| u_{t-1}\!-\!u_*\|^2\!+\!\| u_t\!-\!u_*\|^2].
\end{align*}
Summing up the above inequality for all $t=1,2,\cdots,T$, we obtain
\begin{align*}
2\eta\sum_{t=1}^{T}R(x_t,y_t)\le&\,\| x_0-x_*\|^2_G-\| x_T-x_*\|^2_G+2\eta^2\sigma^2Td_{1}
\\&\!\!+\eta\rho[\| Ax_0\!+\!By_*\!-\!c\|^2\!-\!\| Ax_T\!+\!By_*\!-\!c\|^2\\&\!\!+\| u_0-u_*\|^2-\| u_T-u_*\|^2]\\
\le&\| x_0-x_*\|^2_G+2\eta^2\sigma^2Td_{1}\\
&\!\!+\eta\rho(\| Ax_0+By_*-c\|^2+\| u_0-u_*\|^2).
\end{align*}
The convexity of $R(\cdot,\cdot)$ can be obtained from the convexity of both $f(\cdot)$ and $g(\cdot)$. Let $\tilde{x}\!=\!\frac{1}{T}\sum_{t=1}^{T}x_t , \tilde{y}\!=\!\frac{1}{T}\sum_{t=1}^{T}y_t$. Then
\begin{align*}
2\eta R(\tilde{x},\tilde{y})&\le2\eta\sum_{t=1}^{T}\frac{1}{T}R(x_t,y_t)\\
&\le\frac{1}{T}[\| x_0-x_*\|^2_G+\eta\rho\| Ax_0+By_*-c\|^2\\&\quad+\eta\rho\| u_0-u_*\|^2]+2\eta^2\sigma^2d_{1},\\
R(\tilde{x},\tilde{y})&\le\frac{1}{2\eta T}[\| x_0-x_*\|^2_G+\eta\rho\| Ax_0+By_*-c\|^2\\&\quad+\eta\rho\| u_0-u_*\|^2]+\eta\sigma^2d_{1}.
\end{align*}
Since $Ax_*+By_*-c=0$, $u_0=-\frac{1}{\rho}(A^T)^\dagger\nabla\! f(x_0)$, and $ u_*=-\frac{1}{\rho}(A^T)^\dagger\nabla\! f(x_*)$, then we have
\begin{align*}
&\;\frac{\rho}{2T}\| Ax_0+By_*-c\|^2=\frac{\rho}{2T}\| A(x_0-x_*)\|^2\\=&\;\frac{\rho}{2T}[A(x_0-x_*)]^T[A(x_0-x_*)]=\frac{\rho}{2T}\| x_0-x_*\|^2_{A^T\!A}\\
\leq&\;\frac{\rho}{2T}\| A^T\!A\|_2\cdot\| x_0-x_*\|^2		
\end{align*}
and
\begin{align*}
\frac{\rho}{2T}\| u_0\!-\!u_*\|^2&=\frac{\rho}{2T}\|\!-\frac{1}{\rho}(A^T)^\dagger\nabla\! f(x_0)+\frac{1}{\rho}(A^T)^\dagger\nabla\! f(x_*)\|^2\\
&=\frac{1}{2\rho T}\|\nabla\! f(x_0)-\nabla \! f(x_*)\|^2_{A^{\dagger}(A^{\dagger})^T}\\&\le\frac{1}{2\rho T}\| A^{\dagger}(A^{\dagger})^T\|_2\cdot\|\nabla\! f(x_0)-\nabla\! f(x_*)\|^2\\
&\le\frac{L_f}{2\rho T}\| A^{\dagger}(A^{\dagger})^T\|_2\cdot\| x_0-x_*\|^2.
\end{align*}

Combining the above results, we have
\begin{align*}
\begin{split}
R(\tilde{x},\tilde{y})\le&\;\big(\frac{\| G\|_2}{2\eta T}\!+\!\frac{\rho\| A^T\!A\|_2}{2T}\!+\!\frac{L_f\| A^{\dagger}\!(A^{\dagger})^T\|_2}{2\rho T}\big)\| x_0\!-\!x_*\|^2\\
&+\eta\sigma^2d_{1}.\!\!\!
\end{split}
\end{align*}

The analysis of $utility$ $bound$ is given below. Let $c_1\!=\!\frac{\| G\|_2}{2\eta}\!+\!\frac{\rho\| A^T\!A\|_2}{2}\!+\!\frac{L_f\| A^{\dagger}\!(A^{\dagger})^T\!\|_2}{2\rho}$, and $\bigtriangleup_2(q)\!=\!\frac{c_2}{n}$. Then we have
\begin{equation}\label{equ030}
\begin{split}
R(\tilde{x},\tilde{y})&\le\frac{c_1}{T}\| x_0-x_*\|^2+\frac{c_2^2\alpha T\eta d_{1}}{2n^2\epsilon\mu}\\
&=\frac{c_1}{T}\| x_0-x_*\|^2+O(\frac{\alpha\eta Td_{1}}{n^2\epsilon\mu}).
\end{split}
\end{equation}
Let $\frac{c_1}{T}\| x_0-x_*\|^2=O(\frac{\alpha\eta Td_{1}}{n^2\epsilon\mu})$. Then
\begin{align}\label{equ031}
T=O(\frac{n\sqrt{\epsilon\mu}}{\sqrt{\alpha\eta d_{1}}}).
\end{align}
Substituting the result in (\ref{equ031}) into (\ref{equ030}), we obtain the following results:
\begin{center}
$utility\;\; bound:R(\tilde{x},\tilde{y})\le O(\frac{\sqrt{\alpha\eta d_{1}}}{n\sqrt{\epsilon\mu}})$,\\
$gradient\;\; complexity=n\cdot T=O(\frac{n^2\sqrt{\epsilon\mu}}{\sqrt{\alpha\eta d_{1}}})$
\end{center}
where $\alpha=\frac{\log(1/\delta)}{(1-\mu)\epsilon}+1$.
This completes the proof.
\end{proof}

\section*{Acknowledgments}
We thank all the reviewers for their valuable comments. This work was supported by the National Natural Science Foundation of China (Nos.\ 61876220, 61876221, 61976164, 61836009 and U1701267), the Project supported the Foundation for Innovative Research Groups of the National Natural Science Foundation of China (No.\ 61621005), the Program for Cheung Kong Scholars and Innovative Research Team in University (No.\ IRT\_15R53), the Fund for Foreign Scholars in University Research and Teaching Programs (the 111 Project) (No.\ B07048), the Science Foundation of Xidian University (Nos.\ 10251180018 and 10251180019), the National Science Basic Research Plan in Shaanxi Province of China (Nos.\ 2019JQ-657 and 2020JM-194), and the Key Special Project of China High Resolution Earth Observation System-Young Scholar Innovation Fund.

\bibliographystyle{IEEEtran}
\bibliography{IEEEabrv,references}

\begin{thebibliography}{10}
\providecommand{\url}[1]{#1}
\csname url@samestyle\endcsname
\providecommand{\newblock}{\relax}
\providecommand{\bibinfo}[2]{#2}
\providecommand{\BIBentrySTDinterwordspacing}{\spaceskip=0pt\relax}
\providecommand{\BIBentryALTinterwordstretchfactor}{4}
\providecommand{\BIBentryALTinterwordspacing}{\spaceskip=\fontdimen2\font plus
\BIBentryALTinterwordstretchfactor\fontdimen3\font minus
  \fontdimen4\font\relax}
\providecommand{\BIBforeignlanguage}[2]{{%
\expandafter\ifx\csname l@#1\endcsname\relax
\typeout{** WARNING: IEEEtran.bst: No hyphenation pattern has been}%
\typeout{** loaded for the language `#1'. Using the pattern for}%
\typeout{** the default language instead.}%
\else
\language=\csname l@#1\endcsname
\fi
#2}}
\providecommand{\BIBdecl}{\relax}
\BIBdecl

\bibitem{dwork2006calibrating}
C.~Dwork, F.~McSherry, K.~Nissim, and A.~Smith, ``Calibrating noise to
  sensitivity in private data analysis,'' in \emph{The Third Theory of
  Cryptography Conference}, 2006, pp. 265--284.

\bibitem{Dwork:dp}
C.~Dwork and A.~Roth, ``The algorithmic foundations of differential privacy,''
  \emph{Found. Trends Theor. Comput. Sci.}, vol.~9, no. 3-4, pp. 211--407,
  2014.

\bibitem{gong:dpml}
M.~Gong, Y.~Xie, K.~Pan, K.~Feng, and A.~K. Qin, ``A survey on differentially
  private machine learning,'' \emph{IEEE Comput. Intell. Mag.}, vol.~15, no.~2,
  pp. 49--64, 2020.

\bibitem{mcsherry2009differentially}
F.~McSherry and I.~Mironov, ``Differentially private recommender systems:
  Building privacy into the netflix prize contenders,'' in \emph{Proc. 15th ACM
  SIGKDD Int. Conf. Knowledge Discovery and Data Mining}, 2009, pp. 627--636.

\bibitem{machanavajjhala2011personalized}
A.~Machanavajjhala, A.~Korolova, and A.~D. Sarma, ``Personalized social
  recommendations-accurate or private?'' in \emph{Proc. 37th Int. Conf. Very
  Large Data Bases}, 2011, pp. 440--450.

\bibitem{zhu2013differential}
T.~Zhu, G.~Li, Y.~Ren, W.~Zhou, and P.~Xiong, ``Differential privacy for
  neighborhood-based collaborative filtering,'' in \emph{Proc. IEEE/ACM Int.
  Conf. Advances in Social Networks Analysis and Mining}, 2013, pp. 752--759.

\bibitem{zhu2017privacy}
T.~Zhu, G.~Li, W.~Zhou, and P.~Xiong, ``Privacy preserving for tagging
  recommender systems,'' in \emph{IEEE WIC ACM Int. Conf. Web Intelligence and
  Intelligent Agent Technology}, 2013, pp. 81--88.

\bibitem{lindell2011practical}
Y.~Lindell and E.~Omri, ``A practical application of differential privacy to
  personalized online advertising,'' \emph{IACR Cryptology ePrint Archive},
  vol. 2011, p. 152, 2011.

\bibitem{dankar2012application}
F.~K. Dankar and K.~El~Emam, ``The application of differential privacy to
  health data,'' in \emph{Proc. Joint EDBT/ICDT Workshops}, 2012, pp. 158--166.

\bibitem{chamikara2020privacy}
M.~A.~P. Chamikara, P.~Bertok, I.~Khalil, D.~Liu, and S.~Camtepe, ``Privacy
  preserving face recognition utilizing differential privacy,'' \emph{Computers
  \& Security}, vol.~97, 2020.

\bibitem{othman2014privacy}
A.~Othman and A.~Ross, ``Privacy of facial soft biometrics: Suppressing gender
  but retaining identity,'' in \emph{Proc. European Conf. Computer Vision},
  2014, pp. 682--696.

\bibitem{mcsherry2010differentially}
F.~McSherry and R.~Mahajan, ``Differentially-private network trace analysis,''
  \emph{ACM SIGCOMM Computer Communication Review}, vol.~40, no.~4, pp.
  123--134, 2010.

\bibitem{gotz2011publishing}
M.~Gotz, A.~Machanavajjhala, G.~Wang, X.~Xiao, and J.~Gehrke, ``Publishing
  search logs—a comparative study of privacy guarantees,'' \emph{IEEE Trans.
  Knowl. Data Eng.}, vol.~24, no.~3, pp. 520--532, 2011.

\bibitem{erlingsson2014rappor}
{\'U}.~Erlingsson, V.~Pihur, and A.~Korolova, ``Rappor: Randomized aggregatable
  privacy-preserving ordinal response,'' in \emph{Proc. ACM SIGSAC Conf.
  Computer and Communications Security}, 2014, pp. 1054--1067.

\bibitem{applelearning}
D.~P.~T. Apple, ``Learning with privacy at scale,'' \emph{Technical report,
  Apple}, 2017.

\bibitem{ding2017collecting}
B.~Ding, J.~Kulkarni, and S.~Yekhanin, ``Collecting telemetry data privately,''
  in \emph{Proc. Adv. Neural Inf. Process. Syst.}, 2017, pp. 3571--3580.

\bibitem{abowd2018us}
J.~M. Abowd, ``The us census bureau adopts differential privacy,'' in
  \emph{Proc. 24th ACM SIGKDD Int. Conf. Knowledge Discovery \& Data Mining},
  2018, pp. 2867--2867.

\bibitem{abadi2016deep}
M.~Abadi, A.~Chu, I.~Goodfellow, H.~B. McMahan, I.~Mironov, K.~Talwar, and
  L.~Zhang, ``Deep learning with differential privacy,'' in \emph{Proc. ACM
  SIGSAC Conf. Computer and Communications Security}, 2016, pp. 308--318.

\bibitem{Wang2018Differentially}
D.~Wang, M.~Ye, and J.~Xu, ``Differentially private empirical risk minimization
  revisited: Faster and more general,'' in \emph{Proc. Adv. Neural Inf.
  Process. Syst.}, 2018, pp. 2722--2731.

\bibitem{wang2019efficient}
L.~Wang, B.~Jayaraman, D.~Evans, and Q.~Gu, ``Efficient privacy-preserving
  nonconvex optimization,'' \emph{arXiv: 1910.13659}, 2019.

\bibitem{kim:flasso}
S.~Kim, K.~A. Sohn, and E.~P. Xing, ``A multivariate regression approach to
  association analysis of a quantitative trait network,''
  \emph{Bioinformatics}, vol.~25, pp. 204--212, 2009.

\bibitem{tibshirani:glasso}
R.~J. Tibshirani and J.~Taylor, ``The solution path of the generalized lasso,''
  \emph{Annals of Statistics}, vol.~39, no.~3, pp. 1335--1371, 2011.

\bibitem{ouyang2013stochastic}
H.~Ouyang, N.~He, L.~Tran, and A.~Gray, ``Stochastic alternating direction
  method of multipliers,'' in \emph{Proc. Int. Conf. Machine Learning}, 2013,
  pp. 80--88.

\bibitem{huang2020differentially}
Z.~Huang and Y.~Gong, ``Differentially private {ADMM} for convex distributed
  learning: Improved accuracy via multi-step approximation,'' \emph{arXiv:
  2005.07890v1}, 2020.

\bibitem{huang2019dp}
Z.~Huang, R.~Hu, Y.~Guo, E.~Chan-Tin, and Y.~Gong, ``{DP-ADMM}: {ADMM}-based
  distributed learning with differential privacy,'' \emph{IEEE Trans. Inf.
  Forensics Secur.}, vol.~15, pp. 1002--1012, 2019.

\bibitem{wang:admm}
X.~Wang, H.~Ishii, L.~Du, P.~Cheng, and J.~Chen, ``Privacy-preserving
  distributed machine learning via local randomization and {ADMM}
  perturbation,'' \emph{IEEE Trans. Signal Process.}, vol.~68, pp. 4226--4241,
  2020.

\bibitem{chen2020renyi}
C.~Chen and J.~Lee, ``R{\'e}nyi differentially private {ADMM} for non-smooth
  regularized optimization,'' in \emph{Proc. the Tenth ACM Conf. Data and
  Application Security and Privacy}, 2020, pp. 319--328.

\bibitem{wang2019differential}
P.~Wang and H.~Zhang, ``Differential privacy for sparse classification
  learning,'' \emph{Neurocomputing}, vol. 375, no.~29, pp. 91--101, 2020.

\bibitem{johnson:svrg}
R.~Johnson and T.~Zhang, ``Accelerating stochastic gradient descent using
  predictive variance reduction,'' in \emph{Proc. Adv. Neural Inf. Process.
  Syst.}, 2013, pp. 315--323.

\bibitem{dwork2010boosting}
C.~Dwork, G.~N. Rothblum, and S.~Vadhan, ``Boosting and differential privacy,''
  in \emph{Proc. IEEE the 51st Annual Symposium on Foundations of Computer
  Science}, 2010, pp. 51--60.

\bibitem{yu2019gradient}
D.~Yu, H.~Zhang, W.~Chen, T.-Y. Liu, and J.~Yin, ``Gradient perturbation is
  underrated for differentially private convex optimization,'' in \emph{Proc.
  29th Int. Joint Conf. Artificial Intelligence}, 2020, pp. 3117--3123.

\bibitem{nesterov:fast}
Y.~Nesterov, ``A method of solving a convex programming problem with
  convergence rate ${O}(1/k^2)$,'' \emph{Soviet Math. Doklady}, vol.~27, pp.
  372--376, 1983.

\bibitem{qian1999momentum}
N.~Qian, ``On the momentum term in gradient descent learning algorithms,''
  \emph{Neural networks}, vol.~12, no.~1, pp. 145--151, 1999.

\bibitem{nesterov:co}
Y.~Nesterov, \emph{Introductory Lectures on Convex Optimization: A Basic
  Course}.\hskip 1em plus 0.5em minus 0.4em\relax Boston: Springer US, 2014.

\bibitem{ruder2016overview}
S.~Ruder, ``An overview of gradient descent optimization algorithms,''
  \emph{arXiv: 1609.04747}, 2016.

\bibitem{goldstein2014fast}
T.~Goldstein, B.~O'Donoghue, S.~Setzer, and R.~Baraniuk, ``Fast alternating
  direction optimization methods,'' \emph{SIAM J. Imaging Sciences}, vol.~7,
  no.~3, pp. 1588--1623, 2014.

\bibitem{boyd2011distributed}
S.~Boyd, N.~Parikh, E.~Chu, B.~Peleato, J.~Eckstein \emph{et~al.},
  ``Distributed optimization and statistical learning via the alternating
  direction method of multipliers,'' \emph{Foundations and
  Trends{\textregistered} in Machine learning}, vol.~3, no.~1, pp. 1--122,
  2011.

\bibitem{Li2007t}
N.~Li, T.~Li, and S.~Venkatasubramanian, ``L-closeness: Privacy beyond
  k-anonymity and l-diversity,'' in \emph{IEEE Int. Conf. Data Engineering},
  2007, pp. 24--24.

\bibitem{chaudhuri2011differentially}
K.~Chaudhuri, C.~Monteleoni, and A.~D. Sarwate, ``Differentially private
  empirical risk minimization,'' \emph{Journal of Machine Learning Research},
  vol.~12, pp. 1069--1109, 2011.

\bibitem{Fukuchi2017Differentially}
K.~Fukuchi, Q.~K. Tran, and J.~Sakuma, ``Differentially private empirical risk
  minimization with input perturbation,'' in \emph{Proc. Int. Conf. Discovery
  Science}, 2017, pp. 82--90.

\bibitem{Chen2019Renyi}
C.~Chen, J.~Lee, and D.~Kifer, ``Renyi differentially private erm for smooth
  objectives,'' in \emph{Proc. 22nd Int. Conf. Artificial Intelligence and
  Statistics}, 2019, pp. 2037--2046.

\bibitem{mironov2017renyi}
I.~Mironov, ``R{\'e}nyi differential privacy,'' in \emph{Prof. IEEE 30th
  Computer Security Foundations Symposium}, 2017, pp. 263--275.

\bibitem{zheng2016fast}
S.~Zheng and J.~T. Kwok, ``Fast-and-light stochastic {ADMM},'' in \emph{Proc.
  Int. Joint Conf. Artificial Intelligence}, 2016, pp. 2407--2613.

\bibitem{Donoho:soft}
D.~L. Donoho, ``De-noising by soft-thresholding,'' \emph{IEEE Trans. Inform.
  Theory}, vol.~41, no.~3, pp. 613--627, 1995.

\bibitem{Ouyang2012Stochastic}
H.~Ouyang, N.~He, and A.~Gray, ``Stochastic {ADMM} for nonsmooth
  optimization,'' \emph{arXiv: 1211.0632v2}, 2012.

\bibitem{azadi2014towards}
S.~Azadi and S.~Sra, ``Towards an optimal stochastic alternating direction
  method of multipliers,'' in \emph{Proc. Int. Conf. Machine Learning}, 2014,
  pp. 620--628.

\bibitem{El2008Model}
L.~El~Ghaoui, O.~Banerjee, and A.~D'Aspremont, ``Model selection through sparse
  maximum likelihood estimation for multivariate gaussian or binary data,''
  \emph{Journal of Machine Learning Research}, pp. 485--516, 2008.

\bibitem{liu:asadmm}
Y.~Liu, F.~Shang, H.~Liu, L.~Kong, L.~Jiao, and Z.~Lin, ``Accelerated variance
  reduction stochastic {ADMM} for large-scale machine learning,'' \emph{IEEE
  Trans. Pattern Anal. Mach. Intell.}, 2020.

\bibitem{ding:admm}
J.~Ding, X.~Zhang, M.~Chen, K.~Xue, C.~Zhang, and M.~Pan, ``Differentially
  private robust {ADMM} for distributed machine learning,'' in \emph{Proc. 2019
  IEEE Inte. Conf. Big Data}, 2019, pp. 1302--1311.

\bibitem{bellet2018}
A.~Bellet, R.~Guerraoui, M.~Taziki, and M.~Tommasi, ``Personalized and private
  peer-to-peer machine learning,'' in \emph{Proc. 21st Int. Conf. Artificial
  Intelligence and Statistics}, 2018, pp. 473--481.

\bibitem{chen:admm}
C.~Chen, B.~He, Y.~Ye, and X.~Yuan, ``The direct extension of {ADMM} for
  multi-block convex minimization problems is not necessarily convergent,''
  \emph{Math. Comp.}, vol. 155, pp. 57--79, 2016.

\end{thebibliography}

\end{document}